\documentclass{article}

\usepackage[final]{neurips_2025}

\usepackage{array}
            
\usepackage[utf8]{inputenc} % allow utf-8 input
\usepackage[T1]{fontenc}    % use 8-bit T1 fonts

\usepackage{nicefrac}   
\usepackage{url}            % simple URL typesetting
\usepackage{booktabs}       % professional-quality tables
\usepackage{amsfonts}       % blackboard math symbols
\usepackage{nicefrac}       % compact symbols for 1/2, etc.
\usepackage{microtype}      % microtypography
\usepackage{xcolor}         % colors
\usepackage[colorlinks,
            linkcolor=orange,       %%修改此处为你想要的颜色
            anchorcolor=orange,  %%修改此处为你想要的颜色
            citecolor=brown,        %%修改此处为你想要的颜色，例如修改blue为red
            ]{hyperref}
            
\usepackage{verbatim}
\usepackage{graphicx}
\usepackage{mathrsfs}
\usepackage{amsthm,amsmath,amssymb}
\usepackage{cuted}
\usepackage{xspace}
\usepackage{wasysym}
\usepackage{manfnt}

\usepackage{wrapfig}
\usepackage{bm}
\usepackage{multirow}
\usepackage{multicol}
\usepackage{longtable}
\usepackage{booktabs}
\usepackage{colortbl}  %彩色表格需要加载的宏包
\usepackage{algorithm}
\usepackage{algorithmicx}
\usepackage{algpseudocode}
\usepackage[table]{xcolor} % 先 xcolor 且带 table 选项
\usepackage{colortbl}      % 再 colortbl

\makeatletter
\providecommand\@kludgeins{}%
\providecommand\@makespecialcolbox[1]{#1}%
\makeatother

\title{PPMStereo: Pick-and-Play Memory Construction for Consistent Dynamic Stereo Matching}

\author{
Yun Wang\textsuperscript{1}, Junjie Hu\textsuperscript{2}\thanks{Corresponding author.},   Qiaole Dong\textsuperscript{3}, Yongjian Zhang\textsuperscript{4}\\
\textbf{Yanwei Fu\textsuperscript{3}},
\textbf{Tin Lun Lam\textsuperscript{2}}, \textbf{Dapeng Wu\textsuperscript{1}} \\
\textsuperscript{1}City University of Hong Kong, 
\textsuperscript{2}The Chinese University of Hong Kong, 
Shenzhen\\\textsuperscript{3} Fudan Unvisertiy,\textsuperscript{4} Shenzhen Campus, Sun Yat-sen University \\
{\tt\small ywang3875-c@my.cityu.edu.hk, dpwu@ieee.org,}\\
{\tt\small  \{qldong18, yanweifu\}@fudan.edu.cn}\\
{\tt\small zhangyj85@mail2.sysu.edu.cn,}\\
{\tt\small \{hujunjie,tllam\}@cuhk.edu.cn}
}

\begin{document}
\maketitle

\begin{abstract}
Temporally consistent depth estimation from stereo video is critical for real-world applications such as augmented reality, where inconsistent depth estimation disrupts the immersion of users.
Despite its importance, this task remains challenging due to the difficulty in modeling long-term temporal consistency in a computationally efficient manner.
Previous methods attempt to address this by aggregating spatio-temporal information but face a fundamental trade-off: limited temporal modeling provides only modest gains, whereas capturing long-range dependencies significantly increases computational cost.
To address this limitation, we introduce a memory buffer for modeling long-range spatio-temporal consistency while achieving efficient dynamic stereo matching.
Inspired by the two-stage decision-making process in humans, we propose a \textbf{P}ick-and-\textbf{P}lay \textbf{M}emory (PPM) construction module for dynamic \textbf{Stereo} matching, dubbed as \textbf{PPMStereo}. PPM consists of a `pick' process that identifies the most relevant frames and a `play' process that weights the selected frames adaptively for spatio-temporal aggregation.
This two-stage collaborative process maintains a compact yet highly informative memory buffer while achieving temporally consistent information aggregation.
Extensive experiments validate the effectiveness of PPMStereo, demonstrating state-of-the-art performance in both accuracy and temporal consistency.
% Notably, PPMStereo achieves 0.62/1.11 TEPE on the Sintel clean/final (17.3\% \& 9.02\% improvements over BiDAStereo) with fewer computational costs. 
Codes are available at \textcolor{blue}{https://github.com/cocowy1/PPMStereo}.
\end{abstract}

\section{Introduction}
Stereo matching refers to binocular disparity estimation, which is a fundamental computer vision task focused on estimating the disparity between a pair of rectified stereo images~\cite{wang2024cost,hirschmuller2007stereo,mayer2016large}.
Deep learning-based stereo matching methods have achieved remarkable progress in terms of accuracy~\cite{wang2024cost,xu2023iterative,wang2024selective,cheng2025monster}, efficiency~\cite{tankovich2021hitnet,xu2020aanet,wang2025adstereo,bangunharcana2021correlate}, and robustness~\cite{shen2021cfnet,zhang2024learning,zhang2022revisiting,wang2025learning}.
{Despite impressive performance for static scenes, these methods exhibit severe temporal inconsistencies when applied to dynamic scenes~\cite{karaev2023dynamicstereo}}.
This manifests itself as flickering artifacts and blurred disparity maps due to the absence of effective inter-frame temporal information integration. Therefore, the algorithm deployment in dynamic scenarios such as autonomous driving, robotics, and augmented reality platforms is limited, which requires temporally consistent disparity maps.

\begin{figure*}[t]
    \centering
    \includegraphics[width=0.85\linewidth]{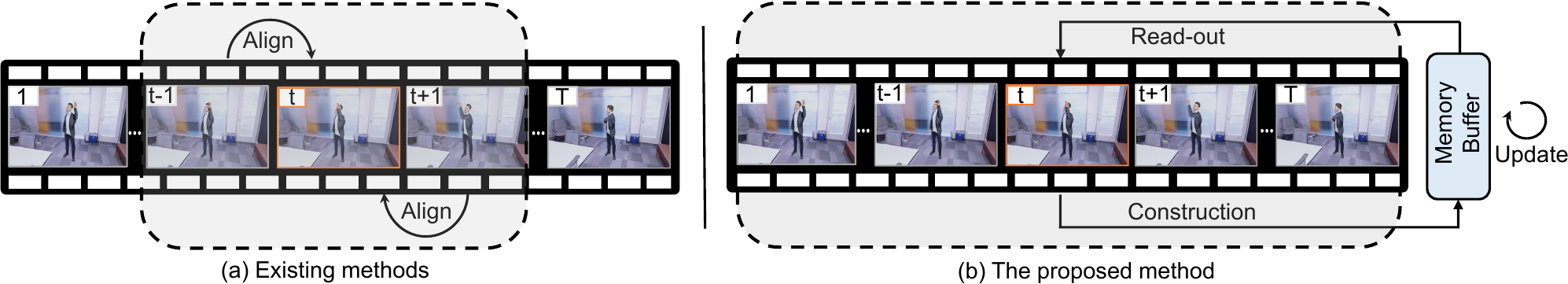}
    \vspace{-0.2cm}
    \caption{Comparison between prior methods (a) and our method (b). For the $t$-th frame, prior works process video sequences using small temporal sliding windows with attention or optical flow, restricting cost information propagation. Our method captures long-range spatio-temporal relationships across the input sequence by constructing and updating a compact memory buffer.}
    \label{sec1:long_range}
    \vspace{-.3cm}
\end{figure*}

To address the task of dynamic stereo matching, {recent approaches start to incorporate temporal cues from two main perspectives to achieve temporally consistent estimation}.
Some methods~\cite{li2023temporally,zeng2024temporally,cheng2024stereo} refine the current disparity {with disparity or motion of previous neighbor frame}, while achieving limited improvements in temporal consistency due to the narrow temporal context.
Secondly, other approaches~\cite{karaev2023dynamicstereo,jing2024match} (Fig.~\ref{sec1:long_range} (a)) expand the temporal receptive field by using attention mechanisms to model spatio-temporal relationships~\cite{karaev2023dynamicstereo} within a sliding window while treating all frames equally, which overlooks variations in frame reliability. 
BiDAStereo~\cite{jing2024match} further depends on optical flow priors for alignment, may incurring errors from flow inaccuracies and high computational cost. Overall, video-based methods face a core trade-off: narrow context yields marginal improvements, whereas naively using all frames drives up computation without reliability awareness.

Naturally, these considerations lead to a key question: How can we design a model that effectively models long-range temporal relationships while maintaining computational efficiency? To answer this question, we draw inspiration from recent advances in sequence processing and bring a memory buffer into the dynamic stereo matching task. We present \textbf{P}ick-and-\textbf{P}lay \textbf{M}emory for dynamic \textbf{Stereo} matching named \textbf{PPMStereo}
which enables effective and efficient utilization of reference frames for long-range spatio-temporal modeling by dynamically reducing redundant frames while selectively retaining and leveraging the most valuable frames throughout the video sequence to ensure accuracy and efficiency, as illustrated in Fig.~\ref{sec1:long_range} (b).

Specifically, our method draws inspiration from human decision-making in complex scenarios, which typically involves the `pick' process that identifies the most essential elements from a set of candidates and the `play' process that meticulously balances and leverages the identified elements~\cite{bhargave2015two,gintis2007framework,santos2015evolutionary}.
In this paper, we propose a novel Pick-and-Play Memory construction method for video stereo matching. Specifically, the `pick' process identifies the most relevant $K$ frames from $T$ reference frames for the current frame. To facilitate this process, we introduce a novel Quality Assessment Module (QAM), which evaluates each frame's contribution by jointly evaluating confidence, redundancy, and similarity of reference frames.
Upon identifying the most relevant $K$ frames, the `play' process adaptively weights the importance of the features extracted from those $K$ selected frames via a dynamic memory modulation mechanism. Subsequently, we utilize an attention-based memory read-out mechanism that queries the high-quality memory buffer using the current frame's contextual feature, yielding temporally and spatially aggregated cost features.
By combining this aggregated cost feature with the current cost
and context features, we can use GRU modules to regress the residual disparities.

Extensive experiments show that our method achieves state-of-the-art temporal consistency and accuracy.
% On both the clean and final pass of the Sintel~\cite{butler2012naturalistic} dataset, our model achieves a temporal end-point error (TEPE) of 0.62 and 1.11 pixels, and 3 pixel error of 5.19\% and 7.64\%, respectively. 
Specifically, on both the clean and final pass of the Sintel~\cite{butler2012naturalistic} dataset, our model achieves a temporal end-of-point error (TEPE) of 0.62 and 1.11 pixels, with 3-pixel error rates of 5.19\% and 7.64\%, respectively. 
Compared to the previous SoTA method, BiDAStereo~\cite{jing2024match}, this represents a 17.3\% and 9.02\% reduction in TEPE and a 9.74\% and 10.32\% improvement in 3-pixel error rate, while enjoying lower computational costs.
% This represents a substantial 17.3\% and 9.02\% error reduction compared to the previous SoTA method BiDAStereo~\cite{jing2024match}, while enjoying lower computational costs.
Overall, the contributions of our work can be summarized as follows:
(1) We introduce PPMStereo, the first work that successfully builds a memory buffer to tackle dynamic stereo matching, allowing for long-range spatio-temporal modeling in a computationally efficient way.
(2) {We propose a novel `Pick-and-Play' memory buffer construction method that first identifies the key subset of reference frames with the pick process and then effectively aggregates them with a play process, enabling highly accurate and temporally consistent disparity estimation.}
(3) Extensive experiments demonstrate that PPMStereo achieves state-of-the-art performance across multiple dynamic stereo matching benchmarks.

\section{Related Work}
\label{sec2}

\textbf{Deep Stereo Matching.}
Existing deep stereo matching methods~\cite{tosi2025survey} primarily focus on cost volume aggregation for network and representation design. These approaches are generally categorized into regression-based~\cite{mayer2016large,kendall2017end,xu2020aanet,zhang2019ga,shen2021cfnet,wang2025dualnet,wang2024cost} and iterative-based methods~\cite{Lipson2021RAFTStereoMR,li2022practical,wang2022spnet,xu2023iterative,wang2024selective,zhang2024learning}.
Regression-based methods typically regress a probability volume to estimate disparity maps, which can be further divided into 2D~\cite{mayer2016large,liang2019stereo,xu2020aanet,wang2022spnet} and 3D cost aggregation approaches~\cite{guo2025lightstereo,zhang2019ga,shen2021cfnet,mao2021uasnet,2023uCFNet,wang2024cost,wang2025adstereo}.
These methods either directly regress disparity across a predefined global range~\cite{kendall2017end,zhang2019ga,wang2024cost} or employ a coarse-to-fine refinement strategy to improve accuracy~\cite{shen2021cfnet,2023uCFNet,mao2021uasnet}.
Recently, iterative-based methods~\cite{wen2025foundationstereo,Lipson2021RAFTStereoMR,bartolomei2025stereo,wang2022spnet,likh2024los,wang2024selective,wang2025learning,wang2025rose,li2025global} have emerged as the dominant paradigm in stereo matching. These methods leverage multi-level GRU or LSTM modules to iteratively refine disparity maps through recurrent cost volume retrieval, achieving state-of-the-art performance.
However, despite their remarkable results, these approaches infer disparities independently for each frame, ignoring temporal correlations across video sequences. As a result, they often suffer from poor temporal consistency, which manifests as flickering artifacts in the disparity outputs.

\textbf{Dynamic Stereo Matching.}
A few methods in stereo matching have focused on leveraging temporal cues from dynamic scenes to enhance disparity consistency.
These methods can be mainly categorized into two paradigms: (i) \textbf{Adjacent-frame Integration}, which propagates disparity or motion fields from the immediately preceding frame to maintain local temporal smoothness. 
These works~\cite{li2023temporally, zhang2023temporalstereo, cheng2024stereo, zeng2024temporally} typically employ warped disparity or motion estimates for robust initialization, thereby enhancing the temporal consistency. However, these methods are limited by their reliance on only the most recent frame, resulting in a narrow temporal receptive field.
(ii) \textbf{Multi-frame Integration}, which employs sliding-window aggregation across extended temporal contexts to enforce temporal consistency through attention mechanisms (DynamicStereo)~\cite{karaev2023dynamicstereo} or optical flow priors (BiDAStereo)~\cite{jing2024match}. Despite their strengths, attention-based methods treat all frames equally without assessing the reliability of reference frames and suffer from high computational costs with a large window. Additionally, flow-based methods are sensitive to optical flow estimation errors and introduce extra computational overheads.
In contrast, our method effectively aggregates long-range spatio-temporal information from a compact yet high-quality memory buffer. {Thanks to our `pick' process, PPMStereo remains computationally efficient, even with the enlarged temporal window. }

\textbf{Memory Cues for Video Tasks.}
Prior works have explored memory model~\cite{sukhbaatar2015end} across various video tasks, including optical flow~\cite{dong2024memflow}, segmentation~\cite{ravi2024sam2,zhou2024rmem,cheng2022xmem,cheng2021rethinking}, tracking~\cite{yang2018learning,fu2021stmtrack}, and video understanding~\cite{song2024moviechat,he2024ma}, demonstrating its significant effectiveness for video-related tasks.
Among them, XMem~\cite{cheng2022xmem} consolidates memory by selecting prototypes and evicting obsolete features via a least-frequently-used policy, while RMem~\cite{zhou2024rmem} improves the segmentation accuracy by using a fixed frame memory bank~\cite{auer2002using}.
Prior works have explored memory model~\cite{sukhbaatar2015end} across various video tasks, including optical flow~\cite{dong2024memflow}, segmentation~\cite{ravi2024sam2,zhou2024rmem,cheng2022xmem,cheng2021rethinking}, and video understanding~\cite{song2024moviechat,he2024ma}, demonstrating its significant effectiveness for video-related tasks.
Among them, XMem~\cite{cheng2022xmem} consolidates memory by selecting prototypes and evicting obsolete features via a least-frequently-used policy, while RMem~\cite{zhou2024rmem} improves the segmentation accuracy by using a fixed frame memory bank~\cite{auer2002using}.
The closest related work is MemFlow~\cite{dong2024memflow}, which develops an adjacent-frame memory buffer framework to aggregate spatio-temporal motion for optical flow estimation. While effective for optical flow, MemFlow yields limited gains when directly applied to dynamic stereo matching, as it only retains the immediate adjacent frame. Expanding its temporal scope without reliability assessment introduces redundant and noisy cues. In contrast, our method adaptively updates and modulates the most valuable memory cues across the entire sequence, enabling robust long-range spatio-temporal modeling while filtering out inferior ones, leading to significant performance improvements.

\section{Methodology}
\label{sec3}

\subsection{Overview}
Dynamic stereo matching seeks to recover a sequence of temporally consistent disparity maps $\left\{\mathbf{d}^t\right\}_{t \in(1, T)} \in \mathbb{R}^{H \times W}$  from stereo video frames $\left\{\mathbf{I}_L^t, \mathbf{I}_R^t\right\}_{t \in(1, T)} \in \mathbb{R}^{H \times W \times 3}$, where $T$ is the number of frames, $H$ and $W$ are the height and width dimensions.
\begin{figure*}[t]
    \centering
    \includegraphics[width=0.95\linewidth]{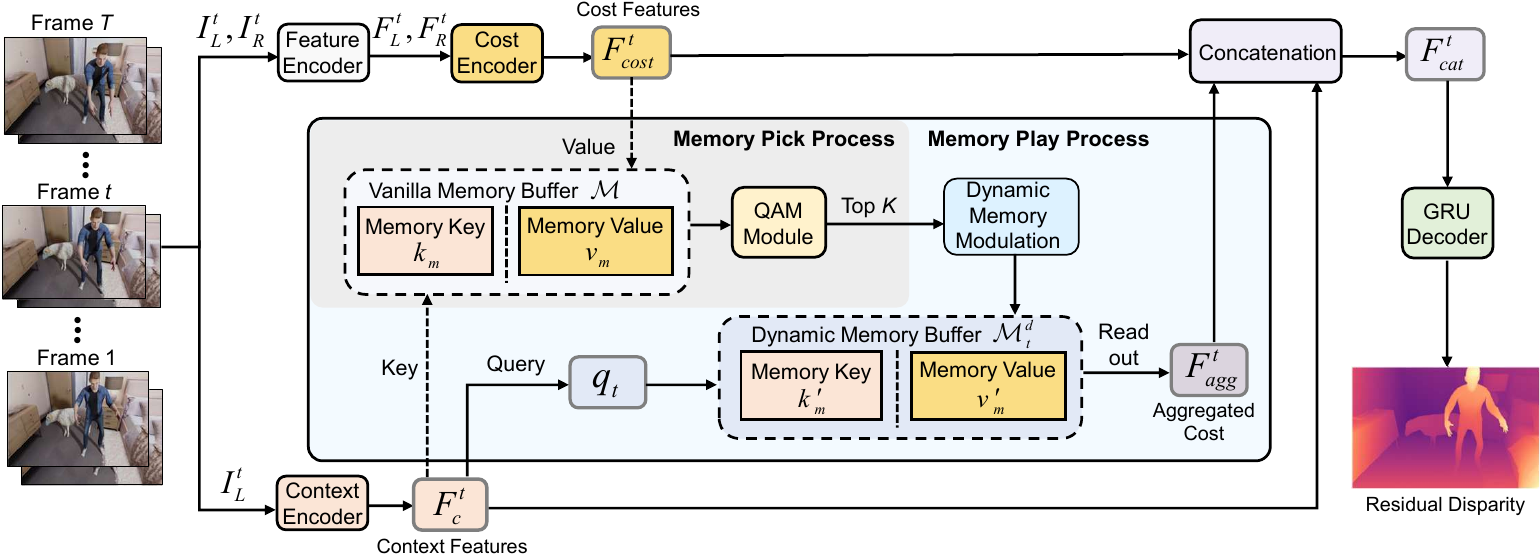}
    \caption{An overview of PPMStereo. The gray part is the memory `pick' process, and the blue part is the memory play process. Our PPMStereo employs a dynamic memory buffer for
modeling long-range spatio-temporal relationships while maintaining computational efficiency.} 
    \label{sec3:overview}
    \vspace{-.3cm}
\end{figure*}
However, prior approaches struggle to capture long-range temporal dependencies without incurring prohibitive cost. To address this, we introduce \textbf{PPMStereo}, which augments the DynamicStereo backbone~\cite{karaev2023dynamicstereo} with a Pick-and-Play Memory (PPM) module that selectively aggregates high-quality references into a compact, query-adaptive buffer, thereby strengthening spatio-temporal modeling while remaining efficient. 
As illustrated in Fig.~\ref{sec3:overview}, the overall pipeline proceeds as follows: 
\textit{(1) Feature Extraction:} a shared encoder extracts multi-scale features $\left\{{F}_L^t, {F}_R^t\right\}_{(s)} \in \mathbb{R}^{sH \times sW \times C}$ at scales $s \in \left\{1/16, 1/8, 1/4 \right\}$, with $C$ channels.
These pyramidal representations provide both receptive-field diversity and a convenient substrate for multi-scale matching.
\textit{(2) Cost Volume Construction:}
at each time step $t$, we construct a 3D correlation volume from $\left\{{F}_L^t, {F}_R^t\right\}_{(s)}$ and pass it through a lightweight cost encoder to obtain matching costs
$F^{t}_{cost}$, subsequently projected to a value embedding $v_{t}$.
\textit{(3) Context Encoding:}
A context encoder operating on the left view produces $F^{t}_{c}$,  which are linearly projected to a query $q_{t}$ and $k_{t}$. 
\textit{(4) Memory Buffer Initialization and Update:} To expose the model to long-range spatio-temporal correlations, we initialize a vanilla memory \(\mathcal{M}=\{\,k_m\in\mathbb{R}^{L\times C},\,v_m\in\mathbb{R}^{L\times C}\,\}\) that stores \(k_m=\{k_1,\ldots,k_T\}\) and \(v_m=\{v_1,\ldots,v_T\}\) with \(L=T\times sH\times sW\). This naive memory buffer stores all reference-frame features, making per-iteration queries prohibitively expensive. To retain accuracy without sacrificing efficiency, we introduce the \emph{Pick-and-Play Memory (PPM)}: driven by a Quality Assessment module (omitting the iteration index \(n\) for brevity), PPM first \emph{picks} the most informative references to construct a compact, dynamic buffer \(\mathcal{M}_t^{d}=\{\,k'_m\in\mathbb{R}^{L'\times C},\,v'_m\in\mathbb{R}^{L'\times C}\,\}\) with \(L'=K\times sH\times sW\) and \(K\ll T\), and then \emph{plays} by adaptively weighting these entries to produce aggregated cost features that balance contributions across the selected frames.
\textit{(6) Iterative Refinement:} following a RAFT-style iterative scheme~\cite{Lipson2021RAFTStereoMR}, we alternate GRU-based updates of disparity estimates with PPM-based memory updates, progressively refining \(\{d_t\}\) while preserving temporal consistency and computational efficiency .

\subsection{Memory Pick Process}

\label{sec3:2}
Naive heuristic strategies, such as random selection or solely keeping the latest frame, are unreliable. Since the former neglects frame reliability and relevance, while the latter suffers from limited temporal context and knowledge drift~\cite{miles2023mobilevos}.
To this end, we introduce a Quality Assessment Module (QAM) that explicitly evaluates the quality of memory elements \(\{k_m, v_m\}\) in the vanilla buffer for dynamic stereo matching. To activate QAM, we define two complementary scores that quantify each reference frame’s contribution to the final accuracy: a confidence score \(\mathbf{S}^c_t\) computed over the value embeddings \(v_m\) to prioritize reliable evidence, and a redundancy-aware relevance score \(\mathbf{S}^r_t\) computed over the key embeddings \(k_m\) to suppress repetitive or low-information entries. The full procedure is summarized in Algorithm~\ref{alg:algorithm1}. \(\mathbf{S}^c_t\) and \(\mathbf{S}^r_t\) are used together to enable the construction of a compact, high-quality memory $\mathcal{M}_{t}^{d}$ that preserves the most informative cross-frame cues.

\begin{algorithm}[t]
\caption{Pseudo code of Pick-and-Play Memory}
\label{alg:algorithm1}
\vspace{0.1cm}
\textbf{Input:} Video frames sequence $\left\{I^{t}_{L}, I^{t}_{R}\right\}$ of video length $T$, GRU $n$-th iterations,  $K$  $\ll$ $T$

\vspace{0.1cm}
\textbf{Intermediates:}  Vanilla Memory: $\mathcal{M}=\left\{k_{m}\in 
 \mathbb{R}^{L\times C}, v_{m}\in\mathbb{R}^{L\times C)}\right\}$,  $L = T\times sH\times sW$ 
 
\qquad \qquad \qquad \; The query: $q_{t}\in\mathbb{R}^{1\times sH\times sW}$, $s\in \left\{1/16, 1/8, 1/4 \right\}$ is the downsampled scale

\qquad \qquad \qquad \; Dynamic Memory: $\mathcal{M}^{d}_{t} =\left\{k'_{m}\in \mathbb{R}^{L'\times C}, v'_{m}\in\mathbb{R}^{L'\times C)}\right\}$, $L' = K\times sH\times sW$

\vspace{0.1cm}
\textbf{Output:} The residual disparity map at $n$-th GRU iteration: $\Delta d^{n}_{t}$ \\
\vspace{0.1cm}

\vspace{0.1cm}
\textbf{1:} \textbf{while} $t \leq T$ \textbf{do}
\vspace{0.1cm}

\vspace{0.1cm}
$\qquad$ \rm{\textbf{Memory Pick Process:}}
\vspace{0.1cm}

\vspace{0.05cm}
\textbf{2:} $\quad$ $\mathbf{S}_{t}=\mathbf{S}^{c}_{t}+\mathbf{S}^{r}_{t}$ $\quad$  \textcolor{blue}{\# QAM, evaluate the  quality of memory elements $k_{m}$ and $v_{m}$}\\

 \textbf{3:} $\quad$ $\mathcal{I}_{t}=\left\{i\mid\operatorname{rank}\left(\mathbf{S}_t[i]\right)\leqslant K\right\}$ \textcolor{blue}{\# Select top-$K$ reference frames}\\
 
\textbf{4:} $\quad$ $\mathcal{M}_t^{d} = \left\{k'_{m}=\operatorname{Cat}\left[\left\{{k}_{i} \mid i \in \mathcal{I}_t\right\}\right], v'_{m}=\operatorname{Cat}\left[\left\{{v}_{i} \mid i \in \mathcal{I}_t\right\}\right]\right\}$ \\

\vspace{-0.1cm}
$\qquad$ \rm{\textbf{Memory Play Process:}}
\vspace{0.15cm}

\textbf{5:}  $\quad$ $\overline{\mathbf{S}}_t[i] = \frac{{\mathbf{S}}_t[i]}{\sum_{i}\mathbf{S}_t[i]}, i\in \mathcal{I}_{t}$ $\quad$ \textcolor{blue}{\# Balance the contribution of selected memory entries}\\

\textbf{6:} $\quad$ $q_{t} = q_{t} + p_{t}$,$\quad$ $k'_{m} = \overline{\mathbf{S}}_t \cdot  k'_{m} + P_{\mathcal{I}_{t}}$ $\quad$ \textcolor{blue}{\# Dynamic memory modulation} \\

\textbf{7:} $\quad$ $F^{t}_{agg}$ = Read-out($q_{t}$, $\mathcal{M}_{t}^{d}$)  $\quad$ 
 \textcolor{blue}{\# Aggregate high-quality spatio-temporal cost information}\\
        
\textbf{8:} $\quad$ $\Delta  d^{n}_{t}=\mathrm{GRU}(F^t_{agg},F^{t}_{cost},F^{t}_{c})$ $\quad$   \textcolor{blue}{\# Produce the disparity map at the $n$-th iteration}\\

\vspace{-0.2cm}
\end{algorithm}

\textbf{Confidence Score.}
Memory values $v_{m}$ encode pixel-wise horizontal displacements, which are critical for disparity estimation.
These features naturally indicate the reliability of its disparity estimation.
To this end, we employ a lightweight confidence network\footnote{The confidence network consists of two convolutional layers followed by a sigmoid activation,  which ensures efficient and effective confidence estimation.} that transforms $v_{m} \in \mathbb{R}^{T \times sH \times sW \times C}$ into confidence maps $u_{t} \in \mathbb{R}^{T \times sH \times sW}$, quantifying whether memory values $v_{m}$ corresponding to accurate disparity outputs.
These confidence maps can provide a frame-level reliability measure by estimating the uncertainty of predicted disparity~\cite{shen2021cfnet,wang2022uncertainty}.
During training for $N$ iterations, the confidence maps are supervised using an $L_1$ loss function to enforce consistency with their ground-truth counterparts. 
The ground-truth confidence score $\hat{u}_{t}$ is computed as follows:
\begin{equation}
\hat{u}_{t}=\exp\left(-\left|\frac{{d}_{t}-\hat{d}_{t}}{\sigma}\right|\right) ,
\end{equation}
where ${d}_{t}$ and $\hat{d}_{t}$ represent the predicted and ground-truth disparities for the $t$-th frame, respectively, and $\sigma$ is a hyper-parameter empirically set to 5. 
{Over $N$ iterations, we compute the confidence loss $L_{conf}$ across all timesteps $u_{t\in(1,T)}$ as follows:}
\begin{equation}
\label{sec3:confidence_loss}
  \mathcal{L}_{conf} = \sum_{t=1}^T \sum_{n=1}^N \gamma^{N-n}\left\|{{u}}_{t}^{n}-\hat{u}_t^{n}\right\|_1,
\end{equation}
where $n$ denotes the number of iterations and $\gamma$ is a decay factor set as 0.9. To obtain a frame-level confidence score $\mathbf{S}^{c}_{t}\in \mathbb{R}^{1\times T}$, we apply average pooling across the spatial dimensions of the confidence maps $u_{t}$.

\textbf{Redundancy-aware Relevance Score.}
Relying solely on the confidence score is insufficient, as adjacent frames often exhibit strong spatio-temporal correlations, which can result in higher confidence scores. This, in turn, introduces feature redundancy and suppresses contributions from more diverse frames, ultimately limiting the diversity and effectiveness of the memory buffer. To mitigate this issue, we propose a redundancy-aware relevance score to evaluate memory keys $k_{m}$, balancing semantic consistency and memory diversity.
First, we compute an inter-frame similarity score $\mathbf{Sim}_{t}\in \mathbb{R}^{1\times T}$ between the current query $q_{t}$ and the memory keys $k_{m}$, measuring semantic alignment while preserving temporal coherence. 
For computational efficiency, we employ an attention mechanism combined with spatial downsampling. Specifically, average pooling reduces the spatial resolution of the query and memory keys from $sH \times sW$ to $sH'\times sW'$, followed by L2-normalization along the combined feature dimension $f=sH' \times sW' \times C$. The similarity score is computed as:
\begin{equation}
\mathbf{sim}_{t} = \phi(q_{t})\phi(k_{m})^{T}, \text{where} \ \ \phi(x)= \frac{{\text{AvgPool}}(x)}{||{\text{AvgPool}}(x)||_{2}}
\end{equation}
where $\phi(k_{m})\in\mathbb{R}^{T\times f}$ and \text{AvgPool($\cdot$)} denotes the average pooling operation.
{However, focusing solely on the most similar regions may overlook occluded areas. Since occluded regions in adjacent frames tend to be highly similar, they can be challenging to reference effectively. To mitigate this,} we then introduce a redundancy-aware regularizer $\mathbf{R}_{t}[k] = e^{-\frac{t_{k}}{T}}$, where $t_{k}$ denotes the the cumulative number of times the $k$-th frame has been selected for the dynamic memory buffer across previous GRU iterations.
{
This term dynamically downweights overused frames while promoting underutilized yet informative references, ensuring a compact yet diverse memory buffer.}  
The final redundancy-aware relevance score $\mathbf{S}_{t}^{r}\in \mathbb{R}^{1\times T}$ combines redundancy and similarity:
\begin{equation}
\mathbf{S}_{t}^{r} = \mathbf{R}_{t}\cdot \mathbf{sim}_{t}
\end{equation}
By jointly considering relevance and diversity, our approach enhances feature aggregation while minimizing redundancy, leading to more robust and efficient memory-based processing.

\begin{figure*}[t]
    \centering
    \includegraphics[width=0.95\linewidth]{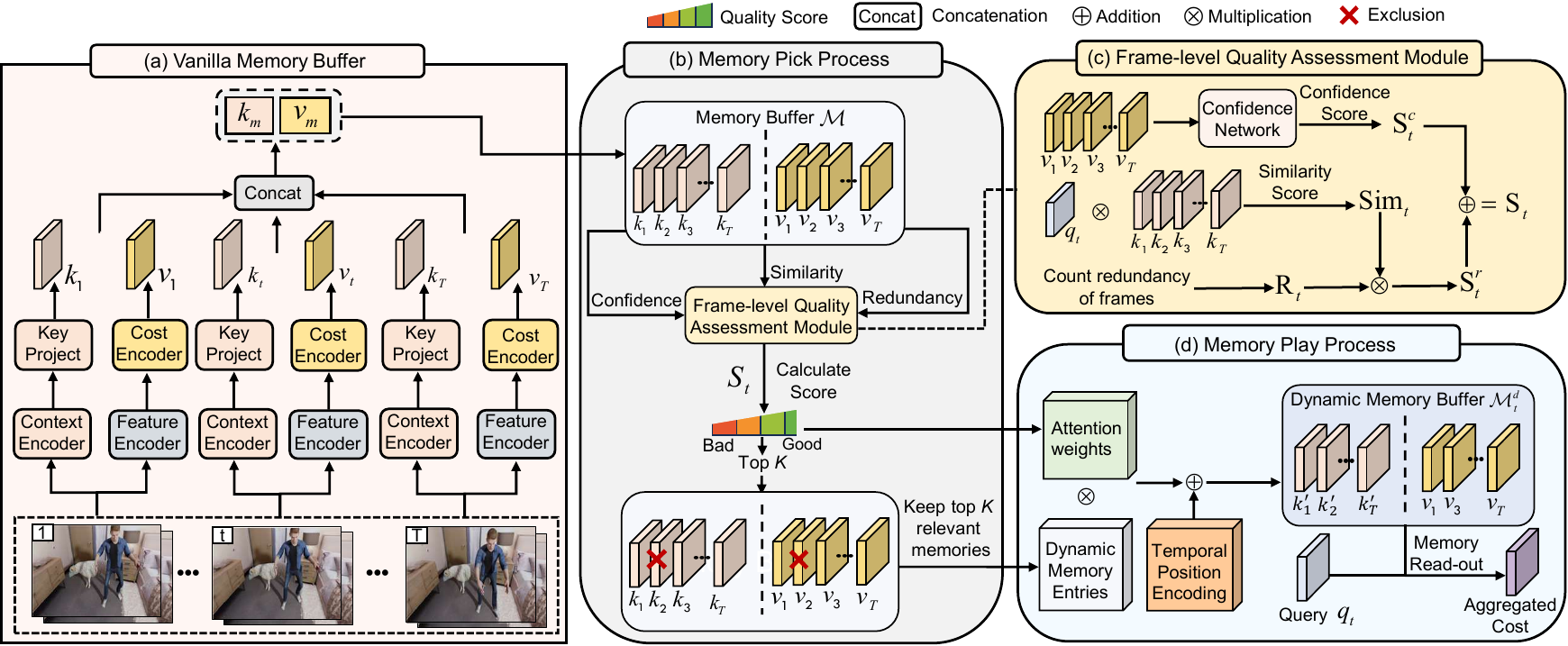}
    \vspace{-0.2cm}
    \caption{The details of our Pick-and-play Memory Construction Process (PPM). } 
    \label{sec3:DMA}
    \vspace{-0.3cm}
\end{figure*}

\textbf{Memory Updating via QAM.}
We compute the total quality metric for each memory frame as  $\mathbf{S}_t = \mathbf{S}^{c}_{t} + \mathbf{S}^{r}_{t}$ by integrating confidence and redundancy-aware relevance scores. This integrated scoring enables dynamic memory update by retaining the most informative entries via a top-$K$ selection mechanism, ensuring robust adaptation to varying video scenarios while preventing memory overload.
{Specifically, for the vanilla memory buffer $\mathcal{M}=\left\{k_{m}, v_{m}\right\}$ with the corresponding quality scores $\mathbf{S}_t\in\mathbb{R}^{1\times T}$, we sort the quality scores in descending order and only retain the top-$K$ memory features in the vanilla memory buffer as:}
% \dql{\begin{equation}
%     % \mathcal{M}^{d}=\left\{k_{i}, v_{i} \mid i \in \mathcal{I}\right\}, \quad 
%     \mathcal{I}=\underset{i \in\{1, \ldots, T\}}{\operatorname{argsort}} \mathbf{S}_i \quad \text { s.t. } \quad|\mathcal{I}|=K,
% \end{equation}
\begin{align}
\mathcal{I}_t & =\left\{i \mid \operatorname{rank}\left(\mathbf{S}_t[i]\right) \leqslant K\right\} \\
\mathcal{M}_t^{d} & = \left\{ \operatorname{Cat}\left[\left\{{k}_{i} \mid i \in \mathcal{I}_t\right\}\right], \operatorname{Cat}\left[\left\{{v}_{i} \mid i \in \mathcal{I}_t\right\}\right]\right\},
\end{align}
\{where rank($\cdot$) denotes the ranking position in descending order, with rank = 1 corresponding to the highest score, $\mathcal{I}_{t}$ is the set of selected frames' indices, and $\operatorname{Cat}$ denotes the concatenation.
The resulting dynamic memory buffer $\mathcal{M}_{t}^d$ comprises keys $k'_{m} = \{k_i\}_{(i\in\mathcal{I}_{t})}$, and values $v'_{m} = \{v_i\}_{(i\in\mathcal{I}_{t})}$.
By enforcing $K\ll T$, this strategy efficiently handles arbitrary video sequences while providing high-quality spatio-temporal cues for dynamic memory aggregation.

\subsection{Memory Play Process}

\label{sec3:3}
{After the pick process selects the top-$K$ most relevant memory entries for our dynamic memory buffer $\mathcal{M}_{t}^{d}$, we argue that not all selected frames contribute equally to disparity estimation. To further weigh their importance, we introduce a memory play process that dynamically weights the selected memory entries based on learned quality scores.} Since dynamic memory construction inherently disrupts temporal ordering, we incorporate temporal position encoding into the framework, ensuring temporal awareness.

\textbf{Dynamic Memory Modulation.}
{Building on this foundation, we propose a unified dynamic memory modulation strategy that jointly optimizes feature reliability and temporal consistency.
Specifically, given the estimated quality score $\mathbf{S}_{t}$, we first obtain the relative significance  of the frames: }
\begin{align}
    \overline{\mathbf{S}}_t[i] = \frac{{\mathbf{S}}_t[i]}{\sum_{i}\mathbf{S}_t[i]}, i\in \mathcal{I}_{t}
\end{align}

Following~\cite{dosovitskiy2020image}, we initialize positional encodings (PE) to align with the original memory buffer length $T$, formalized as $P_{1:T}$. This initialization ensures temporal coherence in feature representation. 
Therefore, the `play' process subsequently operates as follows:
\begin{equation}
   q_{t} = q_{t} + P_{t}, \qquad k'_{m} = \overline{\mathbf{S}}_t \cdot  k'_{m} + P_{\mathcal{I}_{t}}
\end{equation}
where $P_{t}$ denotes the positional encoding at timestep $t$, and $\overline{\mathbf{S}}_t$ represents the aggregated importance weights over the index set $\mathcal{I}_{t}$.
Leveraging the estimated quality scores as reliability indicators, we prioritize more reliable memory entries while maintaining computational efficiency.

\textbf{Memory Read-out.}
We aggregate cost features through an
attention-based memory read-out mechanism from the dynamic memory buffer $\mathcal{M}_{t}^d$.
Specifically, we first compute soft attention weights by measuring the similarity between the query $q_{t}$ and modulated memory keys $k'_{m}$. The aggregated cost features $F_{agg}^{t}$ are then obtained by weighting the memory values $v'_{m}$ through these attention weights:
\begin{equation}
F_{agg}^{t} = F_{cost}^{t} + \alpha \cdot \operatorname{Softmax}\left(1 / \sqrt{D_k} \times q_{t} \times k{'_{m}}^{\mathsf{T}}\right) \times v'_{m},
\end{equation} where $\alpha$ is a learnable scalar initialized from 0. 
In this way,  we employ the attention to gather additional temporal information. 
With the context, cost, and aggregated cost features, we can now output a residual disparity map through a GRU unit at the $n$-th iteration: $\Delta {d}_{n} = \text{GRU}(F_{cost}^{t}, F_{agg}^{t}, F_{c}^{t})$. After $N$ iterations of PPM and GRU, we can get the final disparity map.

\textbf{Loss Functions.}
\label{sec3:losses}
Our disparity loss functions are inherited from the previous works~\cite{karaev2023dynamicstereo,jing2024match}. Generally, for $N$ iterations, we supervise our network with $L_{1}$ distance between our a series of residual flows $\left\{{d}_{1},\ldots, {d}_{T}\right\}$ and the ground-truth $\hat{d}_{t}$ with exponentially increasing weights:
\begin{equation}
\mathcal{L}_{d} = \sum_{t=1}^T \sum_{n=1}^N \gamma^{N-n}\left\|{{d}}^{n}_{t}-\hat{d}_{t}\right\|_1, 
\end{equation}
where $ \gamma$ and $N$ are set as 0.9 and 10, respectively.
Therefore, the total loss function is as follows:
\begin{equation}
\mathcal{L}_{total} = \mathcal{L}_{d} + \mathcal{L}_{conf}. 
\end{equation}

\section{Experiments}
\label{sec4}

\subsection{Datasets}
Our work focuses on videos captured with moving cameras, rendering standard image benchmarks like Middlebury~\cite{middlebury2014}, ETH3D~\cite{Schps2017AMS} unsuitable. For training and evaluation, we employ three synthetic and one real-world stereo video dataset, all featuring dynamic scenes:
\textbf{SceneFlow (SF)}~\cite{mayer2016large} comprising FlyingThings3D, Driving, and Monkaa, with FlyingThings3D featuring moving 3D objects against varied backgrounds.
\textbf{Dynamic Replica (DR)}~\cite{karaev2023dynamicstereo}, a synthetic indoor dataset with non-rigid objects such as people and animals.
\textbf{Sintel~\cite{butler2012naturalistic}}, a synthetic movie dataset available in clean and final passes.
\textbf{South Kensington (SV)}~\cite{jing2024dataset}, a real-world stereo dataset without ground truth data, capturing daily scenarios. We use them for generalization evaluation.
Following prior work~\cite{karaev2023dynamicstereo,jing2024match}, we train on synthetic datasets (SF and DR + SF) and evaluate the performance on \textbf{Sintel}, \textbf{DR}, and \textbf{SV}.

\begin{table}
\scriptsize
\centering
\caption{Quantitative comparison with SoTA methods. Abbreviations: K - KITTI \cite{menze2015object}, M - Middlebury \cite{middlebury2014}, ISV–Infinigen SV~\cite{jing2024dataset}, VK – Virtual KITTI2~\cite{cabon2020virtual}.  CREStereo utilize 7 datasets for training, including SF \cite{mayer2016large}, Sintel \cite{butler2012naturalistic}, FallingThings \cite{tremblay2018falling}, InStereo2K \cite{bao2020instereo2k}, Carla \cite{deschaud2021kitti}, AirSim \cite{shah2018airsim}, and CREStereo dataset \cite{li2022practical}. The best results are in bold, and the second-best are underlined.
% {(accuracy \small{\manconcentriccircles} and temporal consistency \large{\clock}}).
}
\vspace{-0.2cm}
\setlength{\tabcolsep}{1.1pt}
\begin{tabular}
{clc|ccc @{\hspace{10\tabcolsep}} c|ccc @{\hspace{10\tabcolsep}}c|ccc}
\toprule
\multirow{4}{*}{Training data} & \multirow{4}{*}{Method} & \multicolumn{8}{c}{Sintel Stereo} & \multicolumn{4}{c}{Dynamic Replica} \\
\cmidrule(lr){3-10}\cmidrule(lr){11-14}
& & \multicolumn{4}{c}{Clean} & \multicolumn{4}{c}{Final} &  \multicolumn{4}{c}{First 150 frames}  \\
\cmidrule(lr){3-6}\cmidrule(lr){7-10}\cmidrule(lr){11-14}
% & & \small{\manconcentriccircles} & \multicolumn{3}{c}{\large{\clock}} & \small{\manconcentriccircles} & \multicolumn{3}{c}{\large{\clock}} & \small{\manconcentriccircles} & \multicolumn{3}{c}{\large{\clock}} \\
& & \phantom{-}$\delta_{3px}$\phantom{-} & TEPE & $\delta^t_{1px}$ & $\delta^t_{3px}$ & \phantom{-}$\delta_{3px}$\phantom{-} & TEPE  & $\delta^t_{1px}$ & $\delta^t_{3px}$ &  \phantom{-}$\delta_{1px}$\phantom{-} & TEPE  & $\delta^t_{1px}$ & $\delta^t_{3px}$ \\
\midrule
\multirow{7}{*}{SF} &  CODD \cite{li2023temporally}  & 8.68 & 1.44 & 10.8 & 5.65 & 17.46 & 2.32 & 18.56 & 9.79 & 6.59 & 0.105 & 1.04 & 0.42 \\
& RAFT-Stereo \cite{Lipson2021RAFTStereoMR} & 6.12 & 0.92 & 9.33 & 4.51 & 10.40 & 2.10 & 13.69 & 7.08 & 5.51 & 0.145 & 2.03 & 0.65 \\
& DynamicStereo~\cite{karaev2023dynamicstereo} & {6.10} & {0.77} & {8.41} & {3.93} & {8.97} & {1.45} & {11.95} & {5.98} & {3.44} & {0.087} & {0.75} & {0.24} \\
& BiDAStereo~\cite{jing2024match} & {5.94} & {0.73} & {8.29} & {3.79} & {8.78} & {1.26} & {11.65} & {5.53} & 5.17 & 0.103 & 1.11 & 0.40 \\
% & StereoAnyVideo$^\dagger$~\cite{jing2025stereo} & \underline{3.99} & \underline{0.73} & \underline{7.76} & \underline{3.47} & \underline{8.96} & \underline{1.16} & \underline{10.81} & \underline{5.11} & 3.97 & 0.061 & 0.66 & 0.24 \\
& \textbf{PPMStereo (Ours)} & \underline{5.34} & \underline{0.64} & \underline{7.38} & \underline{3.40} & \underline{7.87} & \underline{1.14} & \underline{10.12} & \underline{4.99} & \underline{2.95} & \underline{0.066} & \underline{0.67} & \underline{0.23} \\
& \textbf{PPMStereo\_VDA (Ours)} & \textbf{4.62} & \textbf{0.58} & \textbf{6.89} & \textbf{3.08} & \textbf{7.21} & \textbf{1.04} & \textbf{9.84} & \textbf{4.65} & \textbf{2.37} & \textbf{0.059} & \textbf{0.61} & \textbf{0.22} \\
% \midrule
% 
% & DynamicStereo~\cite{karaev2023dynamicstereo} & 5.77 & {0.76} & {8.46} & {3.93} & {8.68} & {1.42} & {11.93} & {5.92} & 3.32 & {0.075} & {0.68} &  {0.23} \\
% & BiDAStereo~\cite{jing2024match} & 5.75 & \underline{0.75} & \underline{8.03} & \underline{3.76} & \underline{8.52} & \underline{1.22} & \underline{11.04} & \underline{5.30} & \underline{2.81} & \underline{0.062} & \underline{0.61} & \underline{0.22}\\
%  & \textbf{PPMStereo (Ours)} & \textbf{5.05} & \textbf{0.62} & \textbf{7.21} & \textbf{3.29} & \textbf{7.44} & \textbf{1.11} & \textbf{9.98} & \textbf{4.89} & \textbf{2.52} & \textbf{0.056} & \textbf{0.52} & \textbf{0.17} \\
\midrule
SF + M + K &  CODD \cite{li2023temporally} & 9.11 & 1.33 & 12.16 & 6.23 & 11.90 & 2.01 & 16.16 & 8.64 & 10.03 & 0.152 & 2.16 & 0.77 \\
SF + M & RAFT-Stereo \cite{Lipson2021RAFTStereoMR} & 5.86 & 0.85 & 8.79 & 4.13 & {8.47} & 1.63 & 12.40 & 6.23 & 3.46 & 0.114 & 1.34 & 0.41  \\
\phantom{-}7 datasets (incl. Sintel) \phantom{-} & CREStereo \cite{li2022exploring} & {4.58} &  {0.67} & {6.36} & {3.26} & {8.17} & {1.90} & {12.29} & {6.87} & {1.75} & 0.088 & 0.88 & 0.29 \\
% DR + SF + ISV + VK & StereoAnyVideo$^\dagger$~\cite{jing2025stereo} & 5.24 & {0.61} & {6.97} & {3.23} & {7.60} & {1.07} & {9.98} & {4.89} & 3.32 & {0.057} & {0.56} &  {0.20} \\
\midrule
DR + SF &  RAFT-Stereo \cite{Lipson2021RAFTStereoMR} & {5.71} & 0.84 & 9.15 & 4.40 & 9.16 & 2.27 & 13.45 & 7.17 & {1.89} & {0.075} &  0.77 & 0.25  \\
DR + SF & DynamicStereo~\cite{karaev2023dynamicstereo} & 5.77 & {0.76} & {8.46} & {3.93} & {8.68} & {1.42} & {11.93} & {5.92} & 3.32 & {0.075} & {0.68} &  {0.23} \\
DR + SF & BiDAStereo~\cite{jing2024match} & 5.75 & {0.75} & {8.03} & {3.76} & {8.52} & {1.22} & {11.04} & {5.30} & {2.81} & {0.062} & {0.62} & {0.22}\\
DR + SF & \textbf{PPMStereo (Ours)}  & \underline{5.19} & \underline{0.62} & \underline{7.21} & \underline{3.29} & \underline{7.64} & \underline{1.11} & \underline{9.98} & \underline{4.87} & \underline{2.52} & \underline{0.057} & \underline{0.60} & \underline{0.20} \\
DR + SF & \textbf{PPMStereo\_VDA (Ours)} & \textbf{4.47} & \textbf{0.56} & \textbf{6.69} & \textbf{2.97} & \textbf{7.03} & \textbf{1.02} & \textbf{9.65} & \textbf{4.51} & \textbf{1.81} & \textbf{0.052} & \textbf{0.51} & \textbf{0.17} \\
\bottomrule
\end{tabular}
\label{sec4:tab_sota}
\vspace{-0.4cm}
\end{table}

\subsection{Implementation Details}
We implement PPMStereo in PyTorch, training on 8$\times$ A100 GPUs (batch size = 2) using 320$\times$512 crops from 5-frame sequences, evaluated at full resolution with 20-frame sequences. We use AdamW (lr = 0.0003) with one-cycle scheduling, training for 180$k$ iterations ($\approx$ 4.5 days). Data augmentation follows DynamicStereo~\cite{karaev2023dynamicstereo}, including random crops and saturation shifts. For efficient memory read-out, we employ FlashAttention~\cite{dao2022flashattention}.
% Following ~\cite{jing2024match,karaev2023dynamicstereo}, end-point error (EPE) and t-pixel error rate ($\delta_{npx}$) are adopted and also introduce the definition of $\delta_{npx}$ we use temporal end-point-error (TEPE) to measure the variation of the end-point-error over time, where $\delta^{t}_{npx}$ represents the proportion of pixels with TEPE higher than $n$ pixels.
Following prior works~\cite{jing2024match,karaev2023dynamicstereo}, 
we set the number of evaluation iterations $N$ to $20$, while setting $N = 10$ during training.
Besides, we adopt $n$-pixel error rate ($\delta_{npx}$) for accuracy analysis. Additionally, we use the temporal end-point-error (TEPE) to quantify error variation over time, and $\delta^{t}_{npx}$ denotes the percentage of pixels with TEPE exceeding $n$ pixels.
Lower values on metrics indicate greater temporal consistency and disparity estimation accuracy.
{Besides, we replace our original feature extractor with Video Depth Anything (ViT-Small)~\cite{chen2025video}. This PPMStereo\_VDA variant leverages pre-trained representations to further boost performance.}

\begin{figure*}
    \centering
    \includegraphics[width=1\linewidth]{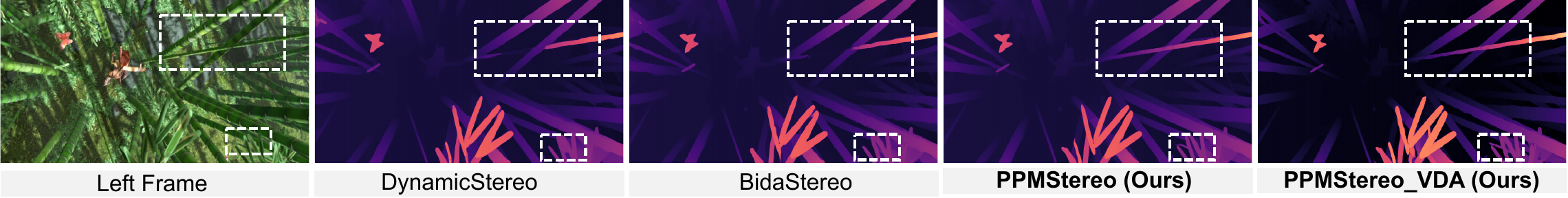}
    \vspace{-0.6cm}    
    \caption{Qualitative comparisons on the Sintel final dataset.} 
    \label{sec4:sintel}
    \vspace{-0.5cm}
\end{figure*}

\subsection{Comparison with State-of-the-Art Methods}

\textbf{Quantitative Results.}
As shown in Tab.~\ref{sec4:tab_sota},
For the SF version, our PPMStereo achieves state-of-the-art performance, outperforming BiDAStereo~\cite{karaev2023dynamicstereo} by 12.3\% \& 9.52\% and DynamicStereo by 16.8\% \& 21.3\% in TEPE on Sintel clean/final pass. The method also demonstrates strong generalization on Dynamic Replica, surpassing all previous approaches across all metrics.
Remarkably, our PPMStereo trained only on synthetic data even largely exceeds the temporal consistency and accuracy of CREStereo~\cite{li2022practical} on Sintel final pass, despite CREStereo using Sintel data for training.
For the SF \& DR version, our method achieves superior temporal consistency with a TEPE of 0.057 on Dynamic Replica, significantly outperforming all previous works. Notably, this is achieved with training on only two synthetic datasets, while CREStereo~\cite{li2022practical} requires seven diverse datasets, demonstrating the efficacy of our long-range temporal modeling.
Overall, the results highlight our method's robust performance and generalization ability in both seen and unseen domains.
{Besides, compared to the previous SoTA method BiDAStereo~\cite{jing2024match}, our method achieves better performance with lower computational costs and memory usage (Please see the appendix for details).}
% As shown in Tab.~\ref{sec4:tab_sota}, our method outperforms prior methods on temporal consistency, achieving 0.057 TEPE on the Dynamic Replica test set.
% Trained on two synthetic datasets, our method shows better temporal performance compared to CREStereo~\cite{li2022practical}, which is trained on 7 diversified
% datasets. This indicates the effectiveness of utilizing temporal information. 
% In Fig.~\ref{sec4:real}, we visualize the rendered images using the disparity predictions of DynamicStereo~\cite{karaev2023dynamicstereo}, BiDAStereo~\cite{jing2024match}, and our PPMStereo to show the superior performance of our approach. The red boxes highlight where our method handles ambiguity better, especially in areas with weak textures such as glass. In contrast, BiDAStereo produces blurry predictions with visible distortions. The results further support our claim that employing local matching and global aggregation via bidirectional alignment can guide the network to enforce temporal consistency.

\begin{wraptable}{r}{0.5\textwidth}
\vspace{-0.65cm}
\caption{Ablations of memory buffer module variants trained on DR+SF.  `OOM' denotes CUDA out of memory.  `Baseline' refers to our backbone model 
 without any memory-related modules.}
% \vspace{-1.2em}
\centering
\setlength{\tabcolsep}{3pt}
\scriptsize
\begin{tabular}{cccc|cc}
\toprule
\multirow{3}{*}{Experiments} & \multirow{3}{*}{Method} & \multicolumn{2}{c}{Sintel Final} & \multicolumn{2}{|c}{Dynamic Replica} \\
\cmidrule(lr){3-6} & & $\delta_{3px}$ & TEPE  & $\delta_{1px}$ & TEPE  \\
\midrule
& Baseline & 8.65 & 1.37  & 3.10 & 0.074 \\
\midrule
\multirow{7}{*}{Memory Buffer} & Full & \multicolumn{2}{c|}{OOM} & \multicolumn{2}{c}{OOM} \\
& MemFlow~\cite{dong2024memflow} & 8.45 & 1.28  & 3.11 & 0.070  \\
& Latest & 8.11 & 1.19 & 2.89 & 0.062  \\
& Random & 8.42 & 1.26 & 2.99 & 0.064  \\
& XMem~\cite{cheng2022xmem} & 8.04 & 1.18 & 2.84 & 0.061 \\
& RMem~\cite{zhou2024rmem} & \underline{7.93} & \underline{1.16} & \underline{2.77} & \underline{0.061}  \\
& Ours & \textbf{7.64} & \textbf{1.11} & \textbf{2.52} & \textbf{0.057} \\
\midrule
\multirow{4}{*}{Memory Length}& $K$ = 1 & 7.95 & 1.18 & 2.70 & 0.062  \\
 & $K$ = 3 & 7.80 & 1.13 & 2.58 & 0.057 \\
  & $K$ = 5 & \underline{7.64} & \underline{1.11}  & \underline{2.52} & \underline{0.057} \\
 & $K$ = 7 & \textbf{7.62} & \textbf{1.10} & \textbf{2.50} & \textbf{0.057} \\
% \multirow{3}{*}{Train Video Length} & 5 & 12.83 & 6.50 & 1.74 & 8.45 & 0.84 & 0.27 & 0.091 & 2.40  \\
% & 7 & 12.83 & 6.50 & 1.74 & 8.45 & 0.84 & 0.27 & 0.091 & 2.40  \\
% & 9 & 12.83 & 6.50 & 1.74 & 8.45 & 0.84 & 0.27 & 0.091 & 2.40  \\
\bottomrule
\end{tabular}
\label{tab:memory_ablation}
\vspace{-0.4cm}
\end{wraptable}

\begin{figure*}
    \centering
    \includegraphics[width=0.98\linewidth]{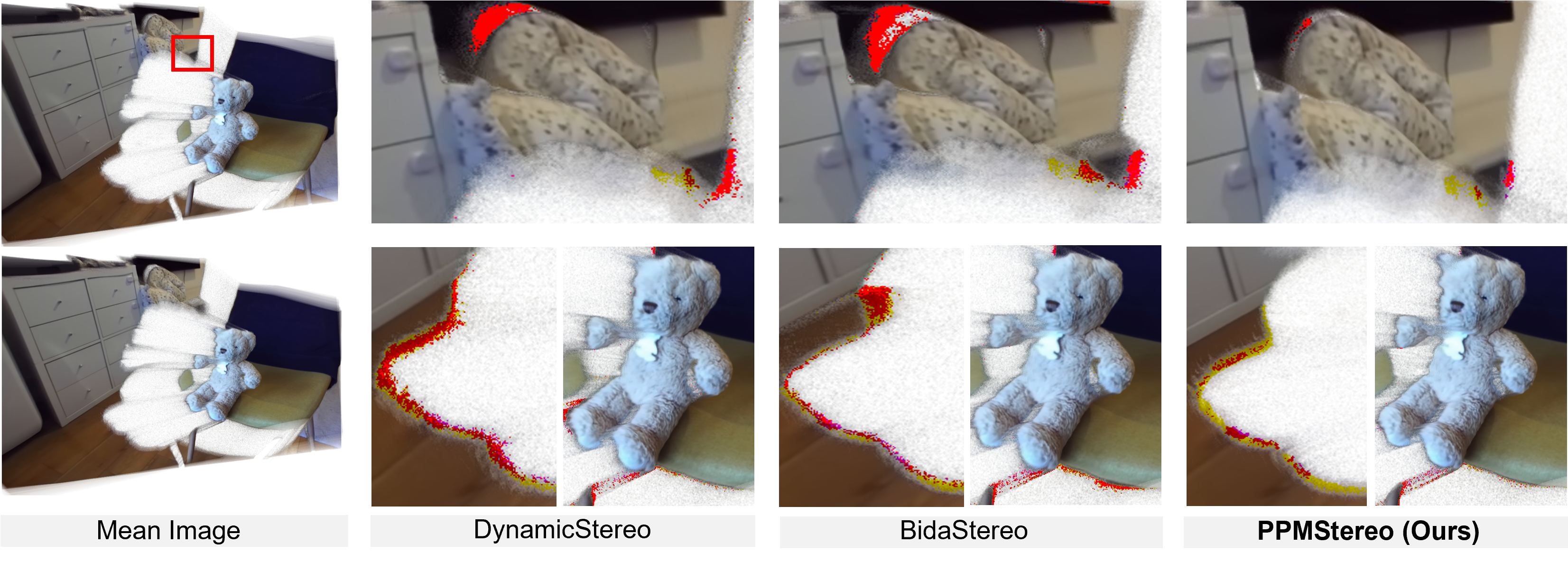}
    \vspace{-0.3cm}    
    \caption{Temporal consistency comparison on 50-frame reconstructed stereo video (all trained on DR + SF). Our method achieves lower variance, demonstrating superior consistency.} 
    \label{sec4:real}
    \vspace{-.1cm}
\end{figure*}

\begin{figure*}
    \centering
    \includegraphics[width=1\linewidth]{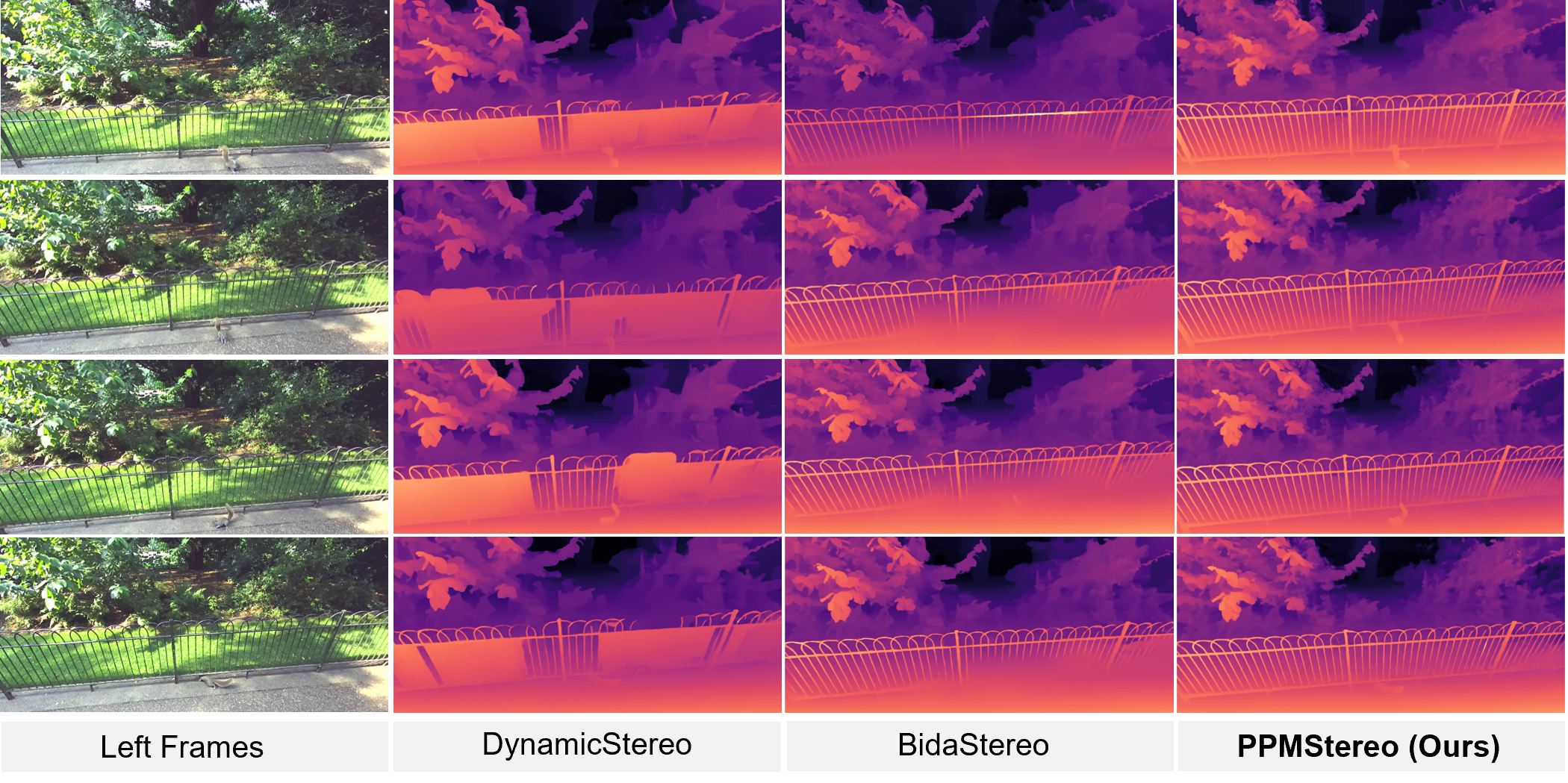}
    \vspace{-0.5cm}    
    \caption{Qualitative generalization comparison on a dynamic outdoor scenario from the SV dataset.} 
    \label{sec4:sk_thin_structure}
    \vspace{-.2cm}
\end{figure*}

\textbf{Qualitative Results.}
Our visual comparisons (Fig.~\ref{sec4:sintel}) using the DR+SF checkpoint show PPMStereo produces sharper disparity predictions than DynamicStereo~\cite{karaev2023dynamicstereo} and BiDAStereo~\cite{jing2024match}, especially in textureless regions (e.g., glass surfaces) where competing methods exhibit blurring artifacts.
Besides, following prior work~\cite{jing2024match,karaev2023dynamicstereo}, we validate temporal consistency on static scenes by rendering depth point clouds at 15-degree viewpoint increments (Fig.~\ref{sec4:real}). Our method shows significantly smaller high-variance regions (> 40 $\textit{px}^{2}$, marked red), confirming superior stability.
Furthermore, on the real-world outdoor scenes from the South Kensington dataset~\cite{jing2024dataset} (Fig.~\ref{sec4:sk_thin_structure}), PPMStereo accurately recovers thin structures such as the fences while maintaining temporal consistency, demonstrating robust generalization to unseen domains. More visualizations are provided in the appendix.

\subsection{Ablation Studies}

% As shown in Tab.~\ref{tab:main_ablation}, we study the specific components of our approach in isolation
% and bold the settings used in the final model. The baseline is based on DynamicStereo~\cite{karaev2023dynamicstereo} with an extra context encoder. 
% We train the models on SceneFlow~\cite{mayer2016large} and Dynamic Replica~\cite{karaev2023dynamicstereo}. 
% We evaluate these models on the final pass of Sintel~\cite{butler2012naturalistic} and on the test split of Dynamic Replica~\cite{karaev2023dynamicstereo}. 
{Due to the huge training cost of PPMStereo\_VPA, we conduct ablation studies exclusively on PPMStereo below. Besides, all ablated models below are trained on DR + SF.}

\textbf{Memory buffer construction.}
We train and evaluate 5 different memory buffer variants, namely, keeping frames from  (1) full frames (20 frames), (2) MemFlow (1 frame)~\cite{dong2024memflow}, (3) the latest frames (5 frames), (4) random (5 frames), (5) XMem~\cite{cheng2022xmem} (distilling all outdated memory features into long-term memory based on attention scores), (6) RMem~\cite{zhou2024rmem} (5 frames), and (7) ours (5 frames).

Specifically, we replace the memory buffer variants and keep the remaining modules unchanged during training and inference. 
Table~\ref{tab:memory_ablation} shows three key insights:
First, while reference frames improve performance, naive accumulation shows diminishing returns, indicating memory capacity alone is insufficient.
Second, frame selection quality critically affects results. The random selection policy underperforms even single-neighbor memory (MemFlow)~\cite{dong2024memflow} on Sintel final pass, highlighting selection importance. However, on the DR dataset with minimal inter-frame changes, the random policy performs comparably to advanced variants.
Lastly, direct long-term memory integration (XMem) shows limited impact, suggesting that simply using all frames may be less effective than the RMem variant.
In contrast, our PPM mechanism overcomes these limitations by dynamically identifying and modulating valuable reference frames, achieving significant TEPE improvements on these two datasets (+19.0\% TEPE on Sintel and +22.9\% TEPE on DR) over the baseline.

\textbf{Memory length.}
Table~\ref{tab:memory_ablation} shows the impact of memory length on PPMStereo. Performance improves initially (e.g., +14.8\% $\delta^{t}_{1px}$ on Sintel  for $K\leq 5$) when trained and evaluated at this memory length, but performance saturates beyond $K$ = 5 due to feature redundancy. To balance computational efficiency and model accuracy, we select $K$ = 5 as the optimal memory length for our final model. 
% This configuration achieves substantial improvements over the baseline ($K=1$) while maintaining efficiency.
% We evaluate the impact of inference memory length on our PPMStereo framework, as summarized in Table~\ref{tab:memory_ablation}. Our experiments reveal that increasing the memory length initially improves performance (e.g., boostingaccuracy by $10.3\%$ for $k\leq 5$), but saturates beyond $k=5$ due to diminishing returns from redundant or outdated features.

\textbf{Contribution of each component.}
% Our analysis (Table~\ref{tab:main_ablation}) 
Table~\ref{tab:main_ablation} shows the proposed PPM module outperforms window-based aggregation through two key processes: (1) The pick process dynamically selects high-quality memory elements from non-adjacent frames, overcoming fixed-window limitations and improving occlusion handling; (2) The play process adaptively weights features by semantic relevance, reducing noise propagation (ID = 3 shows +0.2 on Sintel and +0.017 TEPE improvements on DR compared to the baseline). By combining them, they provide complementary benefits. The pick ensures feature diversity while play suppresses outliers, yielding superior performance in dynamic stereo matching.

\begin{table}
\centering
\scriptsize
\caption{Ablation Study of PPM on Sintel and Dynamic Replica. All models are trained on DR+SF. Note that we directly perform the read-out operation for the ablated model without the `play' process.}
\vspace{-.2cm}
\setlength{\tabcolsep}{9.5pt}
\begin{tabular}{l|cc|cccc|cccc}
\toprule
\multirow{2}{*}{ID} & \multicolumn{2}{c}{Pick-and-Play Memory} & \multicolumn{4}{|c}{Sintel Final} & \multicolumn{4}{|c}{Dynamic Replica} \\
 \cmidrule(lr){2-3}  \cmidrule(lr){4-7}   \cmidrule(lr){8-11}  & Pick & \quad Play   & $\delta_{3px}$  & TEPE & $\delta_{1px}^{t}$ & $\delta_{3px}^{t}$ & $\delta_{1px}$ & TEPE  & $\delta_{1px}^{t}$ & $\delta_{3px}^{t}$ \\
\midrule
1 & \multicolumn{2}{c|}{Baseline} & 8.65 & 1.37 & 11.72 & 5.91 & 3.10 & 0.074 & 0.72 & 0.23  \\
\midrule
% 2 &  & & 11.22 & 5.61 & 1.34 &  8.64 & 0.69 & 0.23 & 0.082 & 3.38  \\
2 & $\checkmark$ & & 7.81  & 1.14 & 10.24 & 5.07 & 2.65 & 0.060  & 0.64  & 0.21 \\
3 & & $\checkmark$ & 7.97 & 1.17 & 10.36 & 5.20 & 2.80 & 0.062 & 0.68 & 0.21 \\
4 & $\checkmark$ & $\checkmark$ & \textbf{7.64} & \textbf{1.11} & \textbf{9.98} & \textbf{4.87}  & \textbf{2.52} & \textbf{0.057}  & \textbf{0.60} & \textbf{0.20} \\
\bottomrule
\end{tabular}
\label{tab:main_ablation}
\vspace{-0.1cm}
\end{table}

\begin{figure*}[h]
    \centering    
    \vspace{-0.1cm}
    \includegraphics[width=0.95\textwidth]{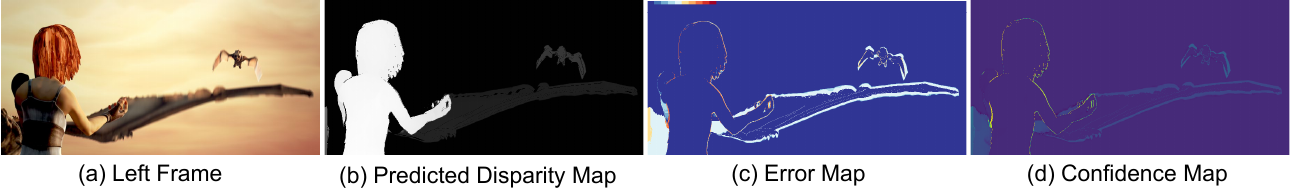}
    \vspace{-0.2cm}
    \caption{Visualization of error map and confidence map. Brighter regions denote higher uncertainty.}
    \vspace{-0.2cm}
    \label{sec4:confidence}
\end{figure*}

\textbf{QAM.}
% Our Quality-Aware Memory (QAM) module dynamically evaluates the reliability of video frames stored in the vanilla memory buffer, leveraging a scoring mechanism detailed in Table~\ref{tab:pick_ablation}. Specifically, we refresh the memory buffer by adaptively balancing two criteria: (1) the quality of stored cost features (memory value $v_{m}$) and (2) the redundancy-aware semantic relevance of frames (memory key $k_{m}$), as described in Sec.~\ref{sec3:2}. As evidenced by Table~\ref{tab:pick_ablation}, integrating the proposed quality score yields consistent performance improvements, achieving depth estimation accuracy and temporal consistency improvements over baseline methods. 
% {Besides, Fig.~\ref{sec4:confidence} shows that the confidence map is highly related to the error map,  demonstrating it as a good quality indicator for $v_{m}$.}
Our QAM module dynamically assesses frame reliability in the memory buffer using a scoring mechanism. 
% (Table~\ref{tab:pick_ablation}). 
We refresh the memory buffer by balancing: (1) cost feature quality ($v_{m}$) and (2) redundancy-aware semantic relevance ($k_{m}$) (Sec.~\ref{sec3:2}). 
Table~\ref{tab:pick_ablation} shows that our quality score improves both depth accuracy and temporal consistency. Fig.~\ref{sec4:confidence} further confirms the confidence map's strong correlation with the error map, validating it as a reliable quality indicator for $v_{m}$.

\begin{table*}
\begin{minipage}{\textwidth}
\begin{minipage}[t]{0.48\textwidth}
\vspace{-0.3cm}
\scriptsize
\caption{Ablation study on the `pick' process. C, Sim, and R denote confidence score, similarity score, and redundancy factor, respectively.}
\setlength{\tabcolsep}{3.2pt}
\scriptsize
\begin{tabular}{l|ccc|ccc|ccc}
\toprule
\multirow{2}{*}{ID} & \multicolumn{3}{c}{QAM} & \multicolumn{3}{|c}{Sintel Final} & \multicolumn{3}{|c}{Dynamic Replica} \\
\cmidrule(lr){2-4} \cmidrule(lr){5-10} & C & Sim & R & $\delta_{3px}$  & TEPE & $\delta_{3px}^{t}$  & $\delta_{1px}$ & TEPE & $\delta_{1px}^{t}$ \\
\midrule
1 & \multicolumn{3}{c|}{Baseline} & 7.97 & 1.17 & 5.20 & 2.80  & 0.062 & 0.68 \\
\midrule
2 & $\checkmark$ & & & 7.81  & 1.14 & 5.06 & 2.63  & 0.058 & 0.65 \\
3 & $\checkmark$ & $\checkmark$ &  &  7.74   & 1.12 & 4.95 & 2.57 & 0.057 & 0.62 \\
4 & $\checkmark$ & $\checkmark$ & $\checkmark$ & \textbf{7.64} & \textbf{1.11} & \textbf{4.87} & \textbf{2.52} & \textbf{0.057} & \textbf{0.60}  \\
\bottomrule
\end{tabular}
\label{tab:pick_ablation}
\end{minipage}\hfill
\begin{minipage}[t]{0.48\textwidth}
\setlength{\tabcolsep}{3pt}
\scriptsize
\centering
\vspace{-0.3cm}
\caption{Ablation study on the `play' process. Weights and PE denote the weighting operation and the temporal position encoding, respectively.}
\begin{tabular}{l|cc|ccc|ccc}
\toprule
\multirow{2}{*}{ID} & \multicolumn{2}{c}{Play Process} & \multicolumn{3}{|c}{Sintel Final} & \multicolumn{3}{|c}{Dynamic Replica} \\
\cmidrule(lr){2-3} \cmidrule(lr){4-9} & Weights & PE & $\delta_{3px}$  & TEPE & $\delta_{3px}^{t}$  & $\delta_{1px}$ & TEPE & $\delta_{1px}^{t}$ \\
\midrule
1 & \multicolumn{2}{c|}{Baseline} &  7.81  & 1.14 & 5.07 & 2.65 & 0.060 & 0.64 \\
\midrule
2 & $\checkmark$ & & 7.67  & 1.12 & 5.00 & 2.54  & 0.060 & 0.62 \\
3 & & $\checkmark$ & 7.77 & 1.11 & 4.93 & 2.63 & 0.058 & 0.61 \\
% & 0.59 & 0.059 & 2.73 \\
4 & $\checkmark$ & $\checkmark$ & \textbf{7.64} & \textbf{1.11} & \textbf{4.87} & \textbf{2.52} & \textbf{0.057} & \textbf{0.60} \\
\bottomrule
\end{tabular}
\label{tab:play_ablation}
\end{minipage}
\end{minipage}
\vspace{-0.5cm}
\end{table*}

\textbf{Memory modulation.}
Our proposed memory modulation mechanism (Sec.~\ref{sec3:3}) further enhances spatio-temporal modeling, achieving a performance gain with +0.17 $\delta_{3px}$ and +0.13 $\delta_{1px}$ improvements {on the Sintel Final and DR, respectively},  as seen in Table~\ref{tab:play_ablation}. 
The adaptive weighting mechanism dynamically prioritizes the most important spatio-temporal features, highlighting accuracy improvements. Meanwhile, learned positional embeddings endow the model with temporal awareness, improving the overall temporal consistency. Experiments show that these components work together to strengthen the model’s ability to capture long-range dependencies and distinguish key spatio-temporal patterns.

\section{Conclusion}
In this paper, we introduce PPMStereo, the first framework, to our knowledge, to leverage high-quality memory for dynamic stereo matching. By selectively updating and modulating the most valuable memory entries, our proposed pick-and-play memory construction mechanism enables the integration of cost information across long-range spatio-temporal connections, ensuring temporally consistent stereo matching. Extensive experiments demonstrate the effectiveness of our approach across diverse datasets, highlighting its generic applicability.

\section*{Acknowledgment}
This work was partly supported by the Shenzhen Science and Technology Program under Grant RCBS20231211090736065, GuangDong Basic and Applied Basic Research Foundation under Grant 2023A151511, Guangdong Natural Science Fund under Grant 2024A1515010252.
This work was also supported by the InnoHK Initiative of the Government of the Hong Kong SAR and the Laboratory for Artificial Intelligence (AI)-Powered Financial Technologies, with additional support from the Hong Kong Research Grants Council (RGC) grant C1042-23GF and the Hong Kong Innovation and Technology Fund (ITF) grant MHP/061/23.
%%%%%%%%%%%%%%%%%%%%%%%%%%%%%%%%%%%%%%%%%%%%%%%%%%%%%%%%%%%%
\bibliography{main}

\begin{thebibliography}{10}

\bibitem{auer2002using}
Peter Auer.
\newblock Using confidence bounds for exploitation-exploration trade-offs.
\newblock {\em Journal of Machine Learning Research (JMLR)}, 3(Nov):397--422, 2002.

\bibitem{bangunharcana2021correlate}
Antyanta Bangunharcana, Jae~Won Cho, Seokju Lee, In~So Kweon, Kyung-Soo Kim, and Soohyun Kim.
\newblock Correlate-and-excite: Real-time stereo matching via guided cost volume excitation.
\newblock In {\em 2021 IEEE International Conference on Intelligent Robots and Systems (IROS)}, pages 3542--3548. IEEE, 2021.

\bibitem{bao2020instereo2k}
Wei Bao, Wei Wang, Yuhua Xu, Yulan Guo, Siyu Hong, and Xiaohu Zhang.
\newblock Instereo2k: a large real dataset for stereo matching in indoor scenes.
\newblock {\em Science China Information Sciences}, 63:1--11, 2020.

\bibitem{bartolomei2025stereo}
Luca Bartolomei, Fabio Tosi, Matteo Poggi, and Stefano Mattoccia.
\newblock Stereo anywhere: Robust zero-shot deep stereo matching even where either stereo or mono fail.
\newblock In {\em Proceedings of the Computer Vision and Pattern Recognition Conference (CVPR)}, pages 1013--1027, 2025.

\bibitem{bhargave2015two}
Rajesh Bhargave, Amitav Chakravarti, and Abhijit Guha.
\newblock Two-stage decisions increase preference for hedonic options.
\newblock {\em Organizational Behavior and Human Decision Processes}, 130:123--135, 2015.

\bibitem{butler2012naturalistic}
Daniel~J Butler, Jonas Wulff, Garrett~B Stanley, and Michael~J Black.
\newblock A naturalistic open source movie for optical flow evaluation.
\newblock In {\em Proceedings of the European conference on computer vision (ECCV)}, pages 611--625. Springer, 2012.

\bibitem{cabon2020virtual}
Yohann Cabon, Naila Murray, and Martin Humenberger.
\newblock Virtual kitti 2.
\newblock {\em arXiv preprint arXiv:2001.10773}, 2020.

\bibitem{chen2025video}
Sili Chen, Hengkai Guo, Shengnan Zhu, Feihu Zhang, Zilong Huang, Jiashi Feng, and Bingyi Kang.
\newblock Video depth anything: Consistent depth estimation for super-long videos.
\newblock 2025.

\bibitem{cheng2022xmem}
Ho~Kei Cheng and Alexander~G Schwing.
\newblock Xmem: Long-term video object segmentation with an atkinson-shiffrin memory model.
\newblock In {\em European Conference on Computer Vision (ECCV)}, pages 640--658. Springer, 2022.

\bibitem{cheng2021rethinking}
Ho~Kei Cheng, Yu-Wing Tai, and Chi-Keung Tang.
\newblock Rethinking space-time networks with improved memory coverage for efficient video object segmentation.
\newblock {\em Advances in Neural Information Processing Systems (NeuralIPS)}, 34:11781--11794, 2021.

\bibitem{cheng2025monster}
Junda Cheng, Longliang Liu, Gangwei Xu, Xianqi Wang, Zhaoxing Zhang, Yong Deng, Jinliang Zang, Yurui Chen, Zhipeng Cai, and Xin Yang.
\newblock Monster: Marry monodepth to stereo unleashes power.
\newblock 2025.

\bibitem{cheng2024stereo}
Ziang Cheng, Jiayu Yang, and Hongdong Li.
\newblock Stereo matching in time: 100+ fps video stereo matching for extended reality.
\newblock In {\em Proceedings of the IEEE Winter Conference on Applications of Computer Vision (WACV)}, pages 8719--8728, 2024.

\bibitem{dao2022flashattention}
Tri Dao, Dan Fu, Stefano Ermon, Atri Rudra, and Christopher R{\'e}.
\newblock Flashattention: Fast and memory-efficient exact attention with io-awareness.
\newblock {\em Advances in Neural Information Processing Systems (NeuralIPS)}, 35:16344--16359, 2022.

\bibitem{deschaud2021kitti}
Jean-Emmanuel Deschaud.
\newblock Kitti-carla: a kitti-like dataset generated by carla simulator.
\newblock {\em arXiv preprint arXiv:2109.00892}, 2021.

\bibitem{dong2024memflow}
Qiaole Dong and Yanwei Fu.
\newblock Memflow: Optical flow estimation and prediction with memory.
\newblock In {\em Proceedings of the IEEE Conference on Computer Vision and Pattern Recognition (CVPR)}, pages 19068--19078, 2024.

\bibitem{dosovitskiy2020image}
Alexey Dosovitskiy, Lucas Beyer, Alexander Kolesnikov, Dirk Weissenborn, Xiaohua Zhai, Thomas Unterthiner, Mostafa Dehghani, Matthias Minderer, Georg Heigold, Sylvain Gelly, et~al.
\newblock An image is worth 16x16 words: Transformers for image recognition at scale.
\newblock {\em arXiv preprint arXiv:2010.11929}, 2020.

\bibitem{fu2021stmtrack}
Zhihong Fu, Qingjie Liu, Zehua Fu, and Yunhong Wang.
\newblock Stmtrack: Template-free visual tracking with space-time memory networks.
\newblock In {\em Proceedings of the IEEE Conference on Computer Vision and Pattern Recognition (CVPR)}, pages 13774--13783, 2021.

\bibitem{gintis2007framework}
Herbert Gintis.
\newblock A framework for the unification of the behavioral sciences.
\newblock {\em Behavioral and brain sciences}, 30(1):1--16, 2007.

\bibitem{guo2025lightstereo}
Xianda Guo, Chenming Zhang, Youmin Zhang, Wenzhao Zheng, Dujun Nie, Matteo Poggi, and Long Chen.
\newblock Lightstereo: Channel boost is all you need for efficient 2d cost aggregation.
\newblock In {\em 2025 IEEE International Conference on Robotics and Automation (ICRA)}, pages 8738--8744. IEEE, 2025.

\bibitem{he2024ma}
Bo~He, Hengduo Li, Young~Kyun Jang, Menglin Jia, Xuefei Cao, Ashish Shah, Abhinav Shrivastava, and Ser-Nam Lim.
\newblock Ma-lmm: Memory-augmented large multimodal model for long-term video understanding.
\newblock In {\em Proceedings of the IEEE Conference on Computer Vision and Pattern Recognition (CVPR)}, pages 13504--13514, 2024.

\bibitem{hirschmuller2007stereo}
Heiko Hirschmuller.
\newblock Stereo processing by semiglobal matching and mutual information.
\newblock {\em IEEE Transactions on Pattern Analysis and Machine Intelligence (TPAMI)}, 30(2):328--341, 2007.

\bibitem{huang2025quaff}
Hong Huang and Dapeng Wu.
\newblock Quaff: Quantized parameter-efficient fine-tuning under outlier spatial stability hypothesis.
\newblock {\em arXiv preprint arXiv:2505.14742}, 2025.

\bibitem{huang2025tequila}
Hong Huang, Decheng Wu, Rui Cen, Guanghua Yu, Zonghang Li, Kai Liu, Jianchen Zhu, Peng Chen, Xue Liu, and Dapeng Wu.
\newblock Tequila: Trapping-free ternary quantization for large language models.
\newblock {\em arXiv preprint arXiv:2509.23809}, 2025.

\bibitem{huang2025fedrts}
Hong Huang, Hai Yang, Yuan Chen, Jiaxun Ye, and Dapeng Wu.
\newblock Fedrts: Federated robust pruning via combinatorial thompson sampling.
\newblock {\em arXiv preprint arXiv:2501.19122}, 2025.

\bibitem{huang2023distributed}
Hong Huang, Lan Zhang, Chaoyue Sun, Ruogu Fang, Xiaoyong Yuan, and Dapeng Wu.
\newblock Distributed pruning towards tiny neural networks in federated learning.
\newblock In {\em 2023 IEEE 43rd International Conference on Distributed Computing Systems (ICDCS)}, pages 190--201. IEEE, 2023.

\bibitem{huang2024fedmef}
Hong Huang, Weiming Zhuang, Chen Chen, and Lingjuan Lyu.
\newblock Fedmef: Towards memory-efficient federated dynamic pruning.
\newblock In {\em Proceedings of the IEEE Conference on Computer Vision and Pattern Recognition (CVPR)}, pages 27548--27557, 2024.

\bibitem{jing2024match}
Junpeng Jing, Ye~Mao, and Krystian Mikolajczyk.
\newblock Match-stereo-videos: Bidirectional alignment for consistent dynamic stereo matching.
\newblock In {\em European Conference on Computer Vision (ECCV)}, pages 415--432. Springer, 2024.

\bibitem{jing2024dataset}
Junpeng Jing, Ye~Mao, Anlan Qiu, and Krystian Mikolajczyk.
\newblock Match stereo videos via bidirectional alignment.
\newblock 2024.

\bibitem{karaev2023dynamicstereo}
Nikita Karaev, Ignacio Rocco, Benjamin Graham, Natalia Neverova, Andrea Vedaldi, and Christian Rupprecht.
\newblock Dynamicstereo: Consistent dynamic depth from stereo videos.
\newblock In {\em Proceedings of the IEEE Conference on Computer Vision and Pattern Recognition (CVPR)}, pages 13229--13239, 2023.

\bibitem{kendall2017end}
Alex Kendall, Hayk Martirosyan, Saumitro Dasgupta, Peter Henry, Ryan Kennedy, Abraham Bachrach, and Adam Bry.
\newblock End-to-end learning of geometry and context for deep stereo regression.
\newblock In {\em Proceedings of the IEEE International Conference on Computer Vision (ICCV)}, pages 66--75, 2017.

\bibitem{li2025global}
Jiahao Li, Xinhong Chen, Zhengmin Jiang, Qian Zhou, Yung-Hui Li, and Jianping Wang.
\newblock Global regulation and excitation via attention tuning for stereo matching.
\newblock {\em arXiv preprint arXiv:2509.15891}, 2025.

\bibitem{li2022practical}
Jiankun Li, Peisen Wang, Pengfei Xiong, Tao Cai, Ziwei Yan, Lei Yang, Jiangyu Liu, Haoqiang Fan, and Shuaicheng Liu.
\newblock Practical stereo matching via cascaded recurrent network with adaptive correlation.
\newblock In {\em Proceedings of the IEEE Conference on Computer Vision and Pattern Recognition (CVPR)}, pages 16263--16272, 2022.

\bibitem{likh2024los}
Kunhong Li, Longguang Wang, Ye~Zhang, Kaiwen Xue, Shunbo Zhou, and Yulan Guo.
\newblock Los: Local structure guided stereo matching.
\newblock In {\em Proceedings of the IEEE Conference on Computer Vision and Pattern Recognition (CVPR)}, 2024.

\bibitem{li2022exploring}
Yanghao Li, Hanzi Mao, Ross Girshick, and Kaiming He.
\newblock Exploring plain vision transformer backbones for object detection.
\newblock In {\em European Conference on Computer Vision (ECCV)}, pages 280--296. Springer, 2022.

\bibitem{li2023temporally}
Zhaoshuo Li, Wei Ye, Dilin Wang, Francis~X Creighton, Russell~H Taylor, Ganesh Venkatesh, and Mathias Unberath.
\newblock Temporally consistent online depth estimation in dynamic scenes.
\newblock In {\em Proceedings of the IEEE Conference on Computer Vision and Pattern Recognition (CVPR)}, pages 3018--3027, 2023.

\bibitem{liang2019stereo}
Zhengfa Liang, Yulan Guo, Yiliu Feng, Wei Chen, Linbo Qiao, Li~Zhou, Jianfeng Zhang, and Hengzhu Liu.
\newblock Stereo matching using multi-level cost volume and multi-scale feature constancy.
\newblock {\em IEEE Transactions on Pattern Analysis and Machine Intelligence (TPAMI)}, 2019.

\bibitem{Lipson2021RAFTStereoMR}
Lahav Lipson, Zachary Teed, and Jia Deng.
\newblock {RAFT}-{S}tereo: Multilevel recurrent field transforms for stereo matching.
\newblock {\em 2021 International Conference on 3D Vision (3DV)}, pages 218--227, 2021.

\bibitem{mao2021uasnet}
Yamin Mao, Zhihua Liu, Weiming Li, Yuchao Dai, Qiang Wang, Yun-Tae Kim, and Hong-Seok Lee.
\newblock Uasnet: Uncertainty adaptive sampling network for deep stereo matching.
\newblock In {\em Proceedings of the IEEE International Conference on Computer Vision (ICCV)}, pages 6311--6319, 2021.

\bibitem{mayer2016large}
Nikolaus Mayer, Eddy Ilg, Philip Hausser, Philipp Fischer, Daniel Cremers, Alexey Dosovitskiy, and Thomas Brox.
\newblock A large dataset to train convolutional networks for disparity, optical flow, and scene flow estimation.
\newblock In {\em Proceedings of the IEEE Conference on Computer Vision and Pattern Recognition (CVPR)}, pages 4040--4048, 2016.

\bibitem{menze2015object}
Moritz Menze and Andreas Geiger.
\newblock Object scene flow for autonomous vehicles.
\newblock In {\em Proceedings of the IEEE Conference on Computer Vision and Pattern Recognition (CVPR)}, pages 3061--3070, 2015.

\bibitem{miles2023mobilevos}
Roy Miles, Mehmet~Kerim Yucel, Bruno Manganelli, and Albert Saa-Garriga.
\newblock Mobilevos: Real-time video object segmentation contrastive learning meets knowledge distillation.
\newblock In {\em Proceedings of the IEEE conference on Computer Vision and Pattern Recognition (CVPR)}, pages 10480--10490, 2023.

\bibitem{ravi2024sam2}
Nikhila Ravi, Valentin Gabeur, Yuan-Ting Hu, Ronghang Hu, Chaitanya Ryali, Tengyu Ma, Haitham Khedr, Roman R{\"a}dle, Chloe Rolland, Laura Gustafson, Eric Mintun, Junting Pan, Kalyan~Vasudev Alwala, Nicolas Carion, Chao-Yuan Wu, Ross Girshick, Piotr Doll{\'a}r, and Christoph Feichtenhofer.
\newblock Sam 2: Segment anything in images and videos.
\newblock {\em arXiv preprint arXiv:2408.00714}, 2024.

\bibitem{santos2015evolutionary}
Laurie~R Santos and Alexandra~G Rosati.
\newblock The evolutionary roots of human decision making.
\newblock {\em Annual review of psychology}, 66(1):321--347, 2015.

\bibitem{middlebury2014}
Daniel Scharstein, Heiko Hirschm{\"u}ller, York Kitajima, Greg Krathwohl, Nera Ne{\v{s}}i{\'c}, Xi~Wang, and Porter Westling.
\newblock High-resolution stereo datasets with subpixel-accurate ground truth.
\newblock In {\em German conference on pattern recognition (GCPR)}, pages 31--42. Springer, 2014.

\bibitem{Schps2017AMS}
Thomas Sch{\"o}ps, Johannes~L. Sch{\"o}nberger, S.~Galliani, Torsten Sattler, Konrad Schindler, Marc Pollefeys, and Andreas Geiger.
\newblock A multi-view stereo benchmark with high-resolution images and multi-camera videos.
\newblock In {\em Proceedings of the IEEE Conference on Computer Vision and Pattern Recognition (CVPR)}, pages 2538--2547, 2017.

\bibitem{shah2018airsim}
Shital Shah, Debadeepta Dey, Chris Lovett, and Ashish Kapoor.
\newblock Airsim: High-fidelity visual and physical simulation for autonomous vehicles.
\newblock In {\em Field and Service Robotics: Results of the 11th International Conference}, pages 621--635. Springer, 2018.

\bibitem{shen2021cfnet}
Zhelun Shen, Yuchao Dai, and Zhibo Rao.
\newblock {CFNet}: Cascade and fused cost volume for robust stereo matching.
\newblock In {\em Proceedings of the IEEE Conference on Computer Vision and Pattern Recognition (CVPR)}, pages 13906--13915, 2021.

\bibitem{2023uCFNet}
Zhelun Shen, Xibin Song, Yuchao Dai, Dingfu Zhou, Zhibo Rao, and Liangjun Zhang.
\newblock Digging into uncertainty-based pseudo-label for robust stereo matching.
\newblock {\em IEEE Transactions on Pattern Analysis and Machine Intelligence (TPAMI)}, 30(2):1--18, 2023.

\bibitem{song2024moviechat}
Enxin Song, Wenhao Chai, Guanhong Wang, Yucheng Zhang, Haoyang Zhou, Feiyang Wu, Haozhe Chi, Xun Guo, Tian Ye, Yanting Zhang, et~al.
\newblock Moviechat: From dense token to sparse memory for long video understanding.
\newblock In {\em Proceedings of the IEEE Conference on Computer Vision and Pattern Recognition (CVPR)}, pages 18221--18232, 2024.

\bibitem{sukhbaatar2015end}
Sainbayar Sukhbaatar, Jason Weston, Rob Fergus, et~al.
\newblock End-to-end memory networks.
\newblock {\em Advances in Neural Information Processing Systems (NeuralIPS)}, 28, 2015.

\bibitem{tankovich2021hitnet}
Vladimir Tankovich, Christian Hane, Yinda Zhang, Adarsh Kowdle, Sean Fanello, and Sofien Bouaziz.
\newblock {HITN}et: Hierarchical iterative tile refinement network for real-time stereo matching.
\newblock In {\em Proceedings of the IEEE Conference on Computer Vision and Pattern Recognition (CVPR)}, pages 14362--14372, 2021.

\bibitem{tosi2025survey}
Fabio Tosi, Luca Bartolomei, and Matteo Poggi.
\newblock A survey on deep stereo matching in the twenties.
\newblock {\em International Journal of Computer Vision (IJCV)}, 133(7):4245--4276, 2025.

\bibitem{tremblay2018falling}
Jonathan Tremblay, Thang To, and Stan Birchfield.
\newblock Falling things: A synthetic dataset for 3d object detection and pose estimation.
\newblock In {\em Proceedings of the IEEE Conference on Computer Vision and Pattern Recognition Workshops (CVPRW)}, pages 2038--2041, 2018.

\bibitem{wang2022uncertainty}
Chen Wang, Xiang Wang, Jiawei Zhang, Liang Zhang, Xiao Bai, Xin Ning, Jun Zhou, and Edwin Hancock.
\newblock Uncertainty estimation for stereo matching based on evidential deep learning.
\newblock {\em Pattern Recognition}, 124:108498, 2022.

\bibitem{wang2025interventional}
Shuguang Wang, Qian Zhou, Kui Wu, Jinghuai Deng, Dapeng Wu, Wei-Bin Lee, and Jianping Wang.
\newblock Interventional root cause analysis of failures in multi-sensor fusion perception systems.
\newblock {\em perception}, 4:5, 2025.

\bibitem{wang2024selective}
Xianqi Wang, Gangwei Xu, Hao Jia, and Xin Yang.
\newblock Selective-stereo: Adaptive frequency information selection for stereo matching.
\newblock {\em arXiv preprint arXiv:2403.00486}, 2024.

\bibitem{wang2025rose}
Yun Wang, Junjie Hu, Junhui Hou, Chenghao Zhang, Renwei Yang, and Dapeng~Oliver* Wu.
\newblock Rose: Robust self-supervised stereo matching under adverse weather conditions.
\newblock {\em IEEE Transactions on Circuits and Systems for Video Technology (TCSVT)}, 2025.

\bibitem{wang2025adstereo}
Yun Wang, Kunhong Li, Longguang Wang, Junjie Hu, Dapeng~Oliver Wu, and Yulan Guo.
\newblock Adstereo: Efficient stereo matching with adaptive downsampling and disparity alignment.
\newblock {\em IEEE Transactions on Image Processing (TIP)}, 2025.

\bibitem{wang2024cost}
Yun Wang, Longguang Wang, Kunhong Li, Yongjian Zhang, Dapeng~Oliver Wu, and Yulan Guo.
\newblock Cost volume aggregation in stereo matching revisited: A disparity classification perspective.
\newblock {\em IEEE Transactions on Image Processing (TIP)}, 2024.

\bibitem{wang2022spnet}
Yun Wang, Longguang Wang, Hanyun Wang, and Yulan Guo.
\newblock {SPN}et: Learning stereo matching with slanted plane aggregation.
\newblock {\em IEEE Robotics and Automation Letters}, 2022.

\bibitem{wang2025learning}
Yun Wang, Longguang Wang, Chenghao Zhang, Yongjian Zhang, Zhanjie Zhang, Ao~Ma, Chenyou Fan, Tin~Lun Lam, and Junjie Hu.
\newblock Learning robust stereo matching in the wild with selective mixture-of-experts.
\newblock {\em arXiv preprint arXiv:2507.04631}, 2025.

\bibitem{wang2025dualnet}
Yun Wang, Jiahao Zheng, Chenghao Zhang, Zhanjie Zhang, Kunhong Li, Yongjian Zhang, and Junjie Hu.
\newblock Dualnet: Robust self-supervised stereo matching with pseudo-label supervision.
\newblock In {\em Proceedings of the AAAI Conference on Artificial Intelligence (AAAI)}, volume~39, pages 8178--8186, 2025.

\bibitem{wen2025foundationstereo}
Bowen Wen, Matthew Trepte, Joseph Aribido, Jan Kautz, Orazio Gallo, and Stan Birchfield.
\newblock Foundationstereo: Zero-shot stereo matching.
\newblock In {\em Proceedings of the Computer Vision and Pattern Recognition Conference (CVPR)}, pages 5249--5260, 2025.

\bibitem{xu2023iterative}
Gangwei Xu, Xianqi Wang, Xiaohuan Ding, and Xin Yang.
\newblock Iterative geometry encoding volume for stereo matching.
\newblock In {\em Proceedings of the IEEE Conference on Computer Vision and Pattern Recognition (CVPR)}, pages 21919--21928, 2023.

\bibitem{xu2020aanet}
Haofei Xu and Juyong Zhang.
\newblock {AAN}et: Adaptive aggregation network for efficient stereo matching.
\newblock In {\em Proceedings of the IEEE Conference on Computer Vision and Pattern Recognition (CVPR)}, pages 1959--1968, 2020.

\bibitem{yang2018learning}
Tianyu Yang and Antoni~B Chan.
\newblock Learning dynamic memory networks for object tracking.
\newblock In {\em Proceedings of the European Conference on Computer Vision (ECCV)}, pages 152--167, 2018.

\bibitem{zeng2024temporally}
Jiaxi Zeng, Chengtang Yao, Yuwei Wu, and Yunde Jia.
\newblock Temporally consistent stereo matching.
\newblock In {\em European Conference on Computer Vision (ECCV)}, pages 341--359. Springer, 2024.

\bibitem{zhang2019ga}
Feihu Zhang, Victor Prisacariu, Ruigang Yang, and Philip~HS Torr.
\newblock {GA}-{N}et: Guided aggregation net for end-to-end stereo matching.
\newblock In {\em Proceedings of the IEEE Conference on Computer Vision and Pattern Recognition (CVPR)}, pages 185--194, 2019.

\bibitem{zhang2022revisiting}
Jiawei Zhang, Xiang Wang, Xiao Bai, Chen Wang, Lei Huang, Yimin Chen, Lin Gu, Jun Zhou, Tatsuya Harada, and Edwin~R Hancock.
\newblock Revisiting domain generalized stereo matching networks from a feature consistency perspective.
\newblock In {\em Proceedings of the IEEE Conference on Computer Vision and Pattern Recognition (CVPR)}, pages 13001--13011, 2022.

\bibitem{zhang2024learning}
Yongjian Zhang, Longguang Wang, Kunhong Li, Yun Wang, and Yulan Guo.
\newblock Learning representations from foundation models for domain generalized stereo matching.
\newblock In {\em European Conference on Computer Vision (ECCV)}, pages 146--162. Springer, 2024.

\bibitem{zhang2023temporalstereo}
Youmin Zhang, Matteo Poggi, and Stefano Mattoccia.
\newblock Temporalstereo: Efficient spatial-temporal stereo matching network.
\newblock In {\em 2023 IEEE/RSJ International Conference on Intelligent Robots and Systems (IROS)}, pages 9528--9535. IEEE, 2023.

\bibitem{zhou2024rmem}
Junbao Zhou, Ziqi Pang, and Yu-Xiong Wang.
\newblock Rmem: Restricted memory banks improve video object segmentation.
\newblock In {\em Proceedings of the IEEE Conference on Computer Vision and Pattern Recognition (CVPR)}, pages 18602--18611, 2024.

\end{thebibliography}

\newpage
\appendix  

\section*{Appendix for PPMStereo}
Our supplementary material provides extensive additional analysis, implementation details, and discussions, organized as follows:
(A) Demonstration Video and More Visualization (Sec.~\ref{sec:a}).
We include a comprehensive demo video (included in demo\_outputs.zip) showcasing: (1) Real-world dynamic scene reconstructions,
(2) Corresponding disparity maps,
(3) Comparative results under varying conditions.
(B) Implementation Details (Sec.~\ref{sec:b}).
We present complete technical specifications for our PPMStereo\_VDA framework, including:
(1) Model architecture: Detailed network configuration.
(2) Datasets: Descriptions of all benchmark datasets used for evaluation.
(3) Algorithmic details: Detailed pseudo-codes.
(4) Computational analysis: Runtime and GPU memory comparisons.
(5) Memory buffer visualization: Evidence of long-range relationship modeling.
(C) Additional discussions on limitations and future work.
We offer a more detailed discussion of the limitations
and potential future directions (Sec.~\ref{sec:d}).

% Our appendix covers additional analysis, implementation details, and discussion as below:
% (A) Demo Video. We provide a demo video in the named demo\_outputs zip file showing multiple real-world dynamic reconstruction results and disparity maps.
% (B) Implementation Details. We provide more details about our PPMStereo\_VDA model, datasets, Pseudo codes, and computational cost comparison. We also provide a visualization map to provide our memory buffer can introduce long-range relationships.
% (C) Addition discussions on limitations and future work.
% We offer a more detailed discussion of the limitations
% and potential future directions (Sec.~\ref{sec:d}).

\begin{figure*}[h]
    \centering
    \includegraphics[width=1\linewidth]{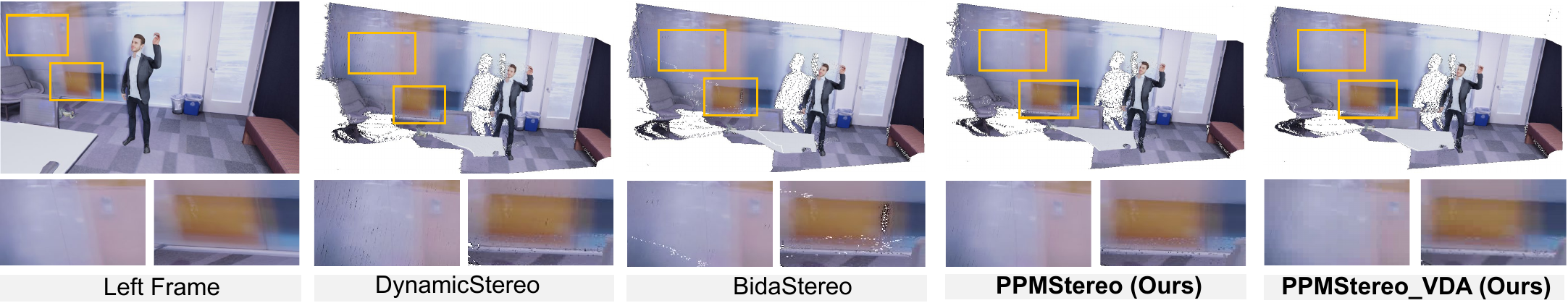} 
    \caption{Qualitative comparisons on the Dynamic Replica test set. They are rendered with a camera displaced by 15 degree angles. Our method exhibits smoother reconstruction results.} 
    \label{sec4:vis_dr}
\end{figure*}

\section{More Visualizations on Real-world Scenes}
\begin{figure*}[h]
    \centering
    \includegraphics[width=1\linewidth]{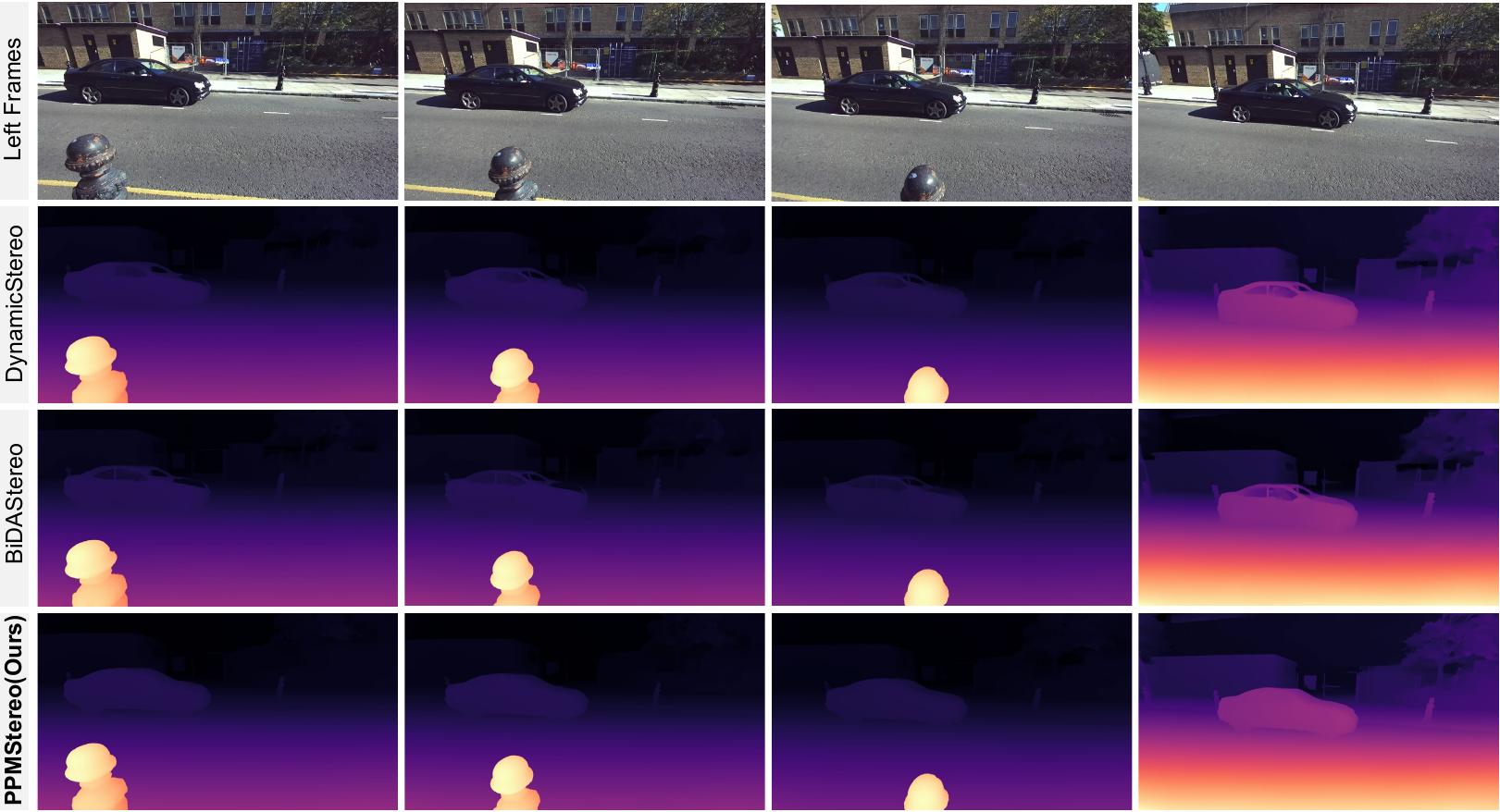} 
    \caption{Qualitative comparison on a dynamic outdoor scenario from the South Kensington SV dataset~\cite{jing2024dataset}.} 
    \label{sec4:sk_outdoor}
\end{figure*}

\begin{figure*}[h]
    \centering
    \includegraphics[width=1\linewidth]{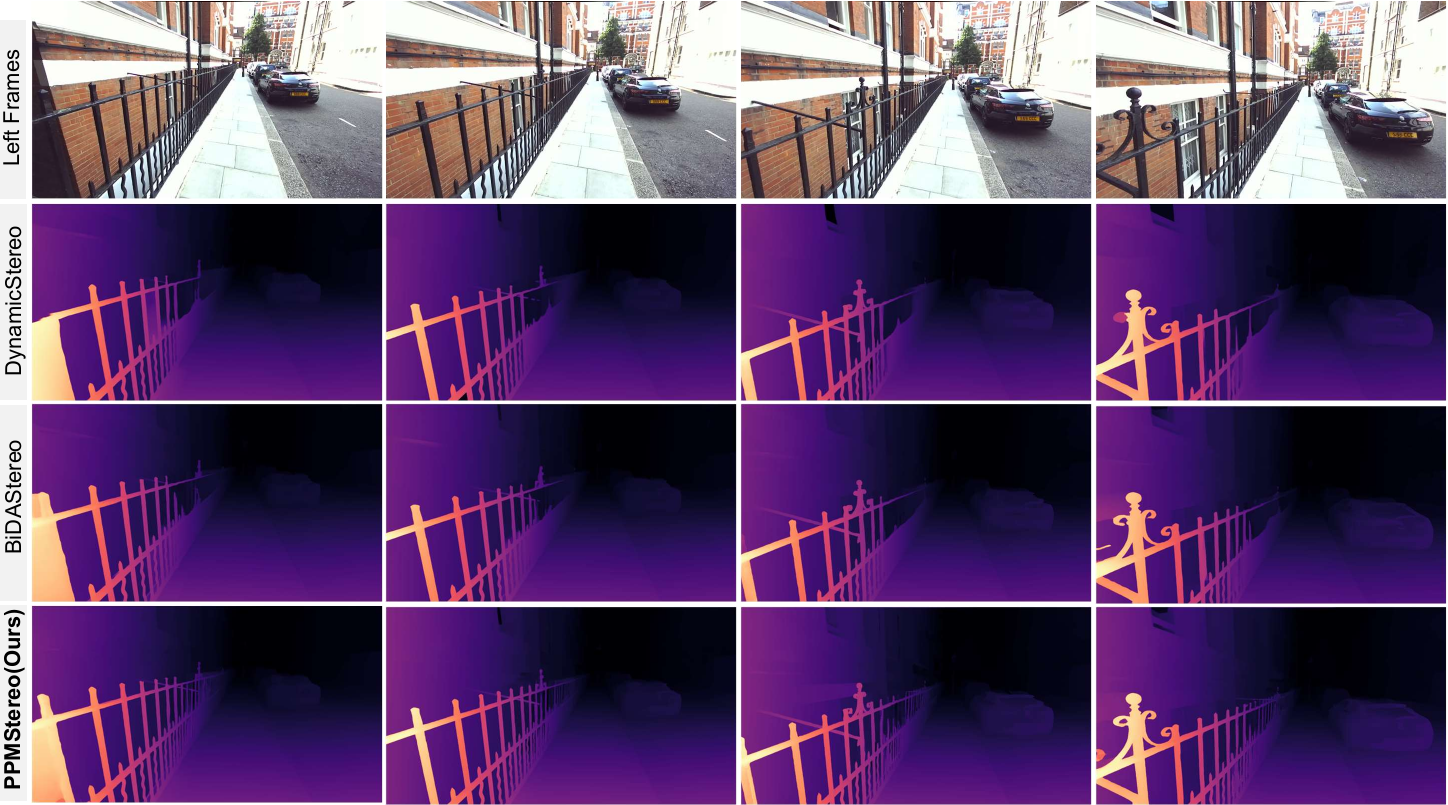}
    \caption{Qualitative comparison on a dynamic outdoor scenario from the South Kensington SV dataset~\cite{jing2024dataset}.} 
    \label{sec4:sk_outdoor3}
\end{figure*}

\begin{figure*}[h]
    \centering
    \includegraphics[width=1\linewidth]{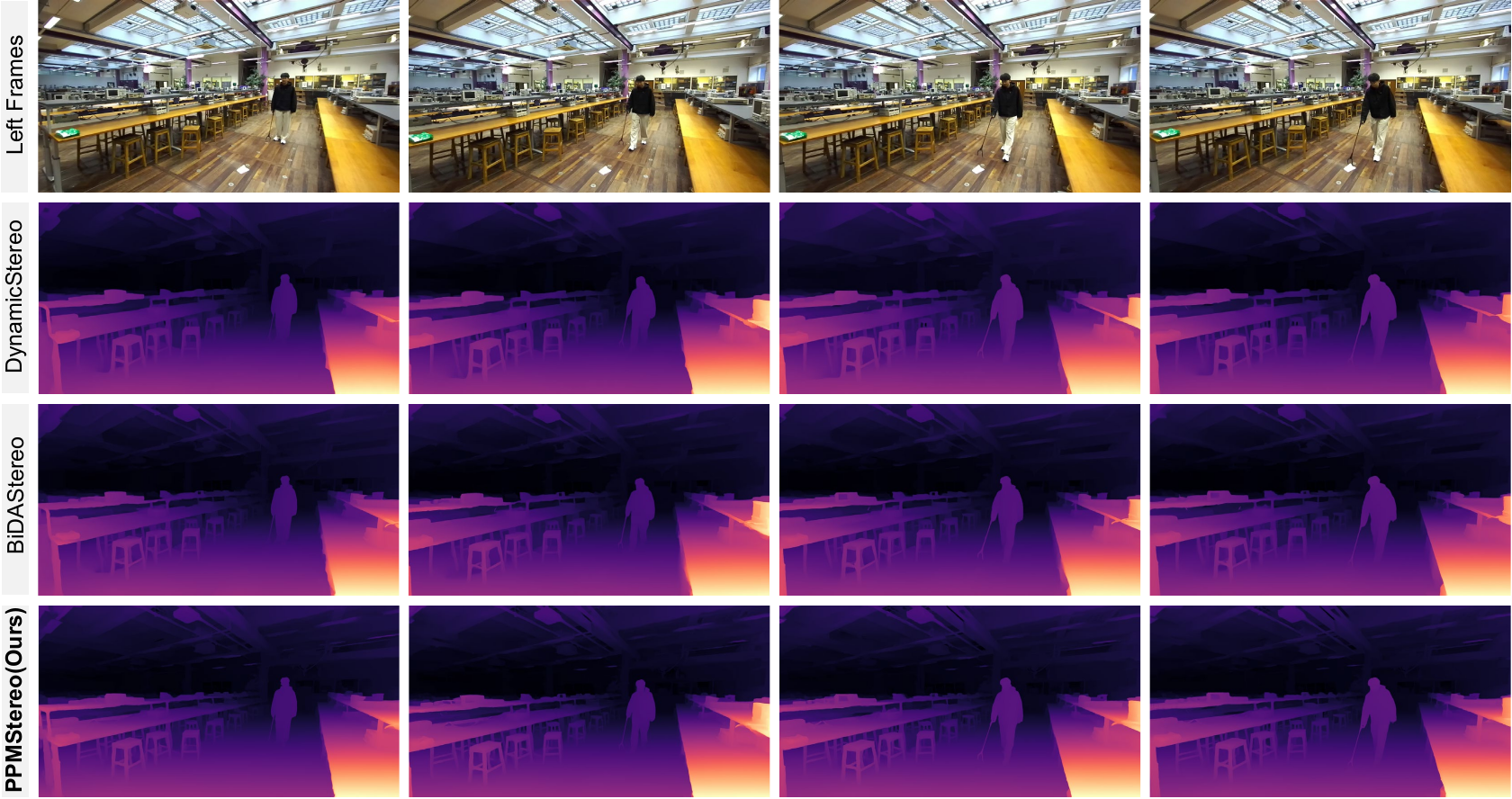}  
    \caption{Qualitative comparison on a dynamic indoor scenario from the South Kensington SV dataset~\cite{jing2024dataset}.} 
    \label{sec4:sk_indoor}
    \vspace{-.3cm}
\end{figure*}

\begin{figure*}[t]
    \centering
    \includegraphics[width=1\linewidth]{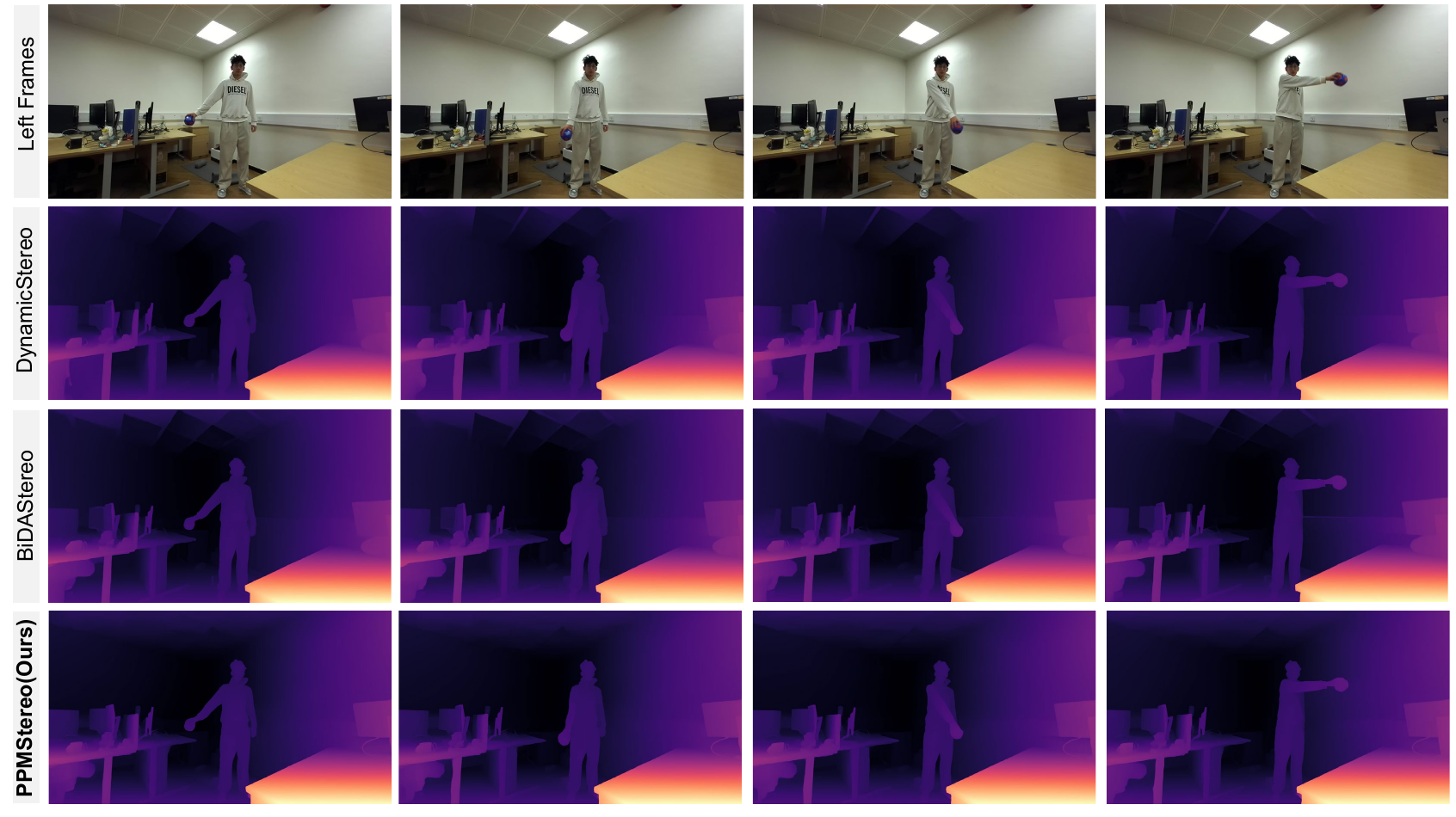}  
    \caption{Qualitative comparison on a dynamic indoor scenario from the South Kensington SV dataset~\cite{jing2024dataset}.} 
    \label{sec4:sk_indoor2}
\end{figure*}

Figure~\ref{sec4:vis_dr} demonstrates the reconstruction performance of our method on the Dynamic Replica (DR) test set. The results illustrate our approach's ability to accurately recover fine-grained details while preserving the global structural integrity of the scene, even under challenging dynamic conditions.
Figure~\ref{sec4:sk_outdoor} and Figure~\ref{sec4:sk_outdoor3} showcase the performance of our method in outdoor real-world scenarios, highlighting its robustness under varying lighting conditions and complex backgrounds. For indoor environments, Figure~\ref{sec4:sk_indoor} and Figure~\ref{sec4:sk_indoor2} provide a comprehensive comparison, demonstrating consistent accuracy even in confined spaces with occlusions and dynamic objects. Additional qualitative results (e.g., thin structures and reconstructed results) are available in the supplementary materials (demo\_outputs.zip).

~\label{sec:a}

\begin{figure*}[t]
    \centering
    \includegraphics[width=1\linewidth]{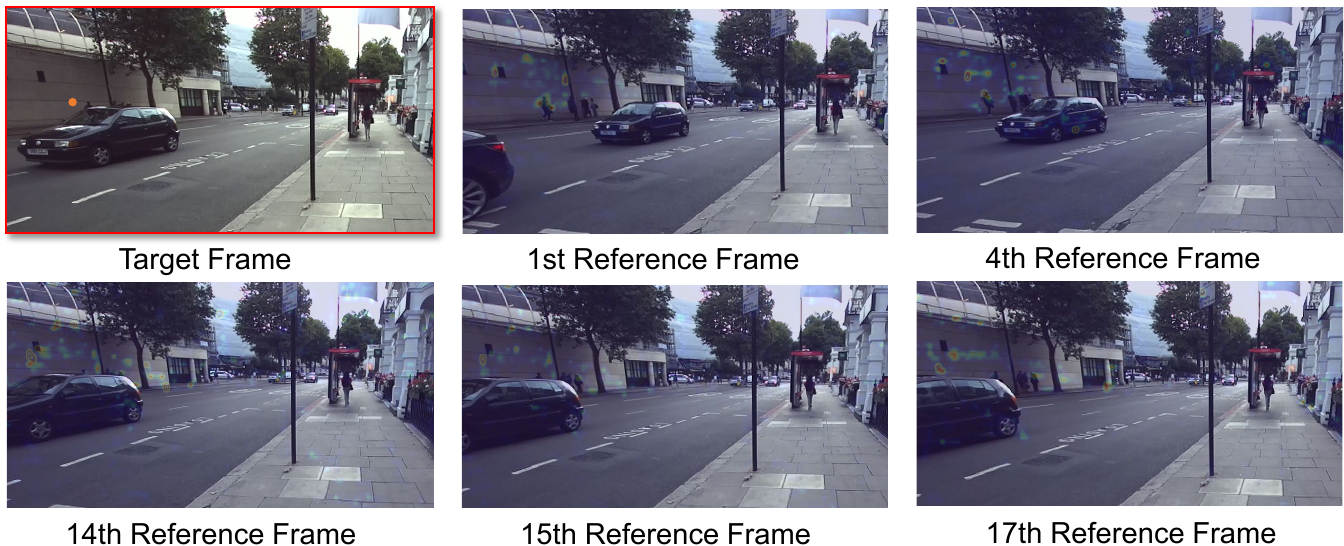}
    \caption{For the target frame (11th frame), the occlusion point is highlighted by a yellow circle. Unlike conventional approaches that rely on adjacent frames, our PPMStereo method dynamically selects and aggregates features from the most informative and diverse frames across the entire sequence ($T$=20). By adaptively bypassing occluded or unreliable neighboring frames, PPMStereo ensures robust and occlusion-aware feature representation, enhancing both accuracy and generalization.}
    \label{sec4:attention_vis}
\end{figure*}

\section{Implementation Details}
~\label{sec:b}
\subsection{PPMStereo\_VDA}
For PPMStereo\_VDA model, we use VideoDepthAnything~\cite{chen2025video}
to replace our feature extractor.
Specifically, in the feature extraction stage, when processing a video sequence with the monocular video depth model, we first resize it to ensure its dimensions are divisible by 14, maintaining consistency with the model’s pretrained patch size. After obtaining the feature maps, we resize the image back to its original dimensions.  
The monocular depth model produces feature maps with 64 channels, while the CNN encoders extract both image and context features with 128 channels each. These feature maps are concatenated to form a 192-channel representation, a decoder is used then 
to obtain a 128-channel representation, which serves as input to the subsequent correlation module.

\subsection{Datasets.}
\textbf{SceneFlow (SF)}
SceneFlow \cite{mayer2016large} consists of three subsets: {FlyingThings3D}, {Driving}, and {Monkaa}.
\begin{itemize}
    \item {FlyingThings3D} is an abstract dataset featuring moving shapes against colorful backgrounds. It contains 2,250 sequences, each spanning 10 frames.
    \item {Driving} includes 16 sequences depicting driving scenarios, with each sequence containing between 300 and 800 frames.
    \item {Monkaa} comprises 48 sequences set in cartoon-like environments, with frame counts ranging from 91 to 501.
\end{itemize}

\textbf{Sintel}
Sintel \cite{butler2012naturalistic} is generated from computer-animated films. It consists of 23 sequences available in both {clean} and {final} rendering passes. Each sequence contains 20 to 50 frames. We use the full sequences of Sintel for evaluation.

\textbf{Dynamic Replica}
Dynamic Replica \cite{karaev2023dynamicstereo} is designed for longer sequences and the presence of non-rigid objects such as animals and humans. The dataset includes:
\begin{itemize}
    \item {484 training sequences}, each with 300 frames.
    \item {20 validation sequences}, each with 300 frames.
    \item {20 test sequences}, each with 900 frames.
\end{itemize}
Following prior methods \cite{karaev2023dynamicstereo, jing2024match}, we use the entire training set for model training and evaluate on the first 150 frames of the test set.

\textbf{South Kensington SV}
South Kensington SV \cite{jing2024match} is a real-world stereo dataset capturing {daily life scenarios} for qualitative evaluation. It consists of {264 stereo videos}, each lasting between {10 and 70 seconds}, recorded at {1280×720 resolution} and {30 fps}. We conduct qualitative evaluations on this dataset.

\subsection{Computational Costs}
\begin{figure*}[t]
    \centering
    \includegraphics[width=1\linewidth]{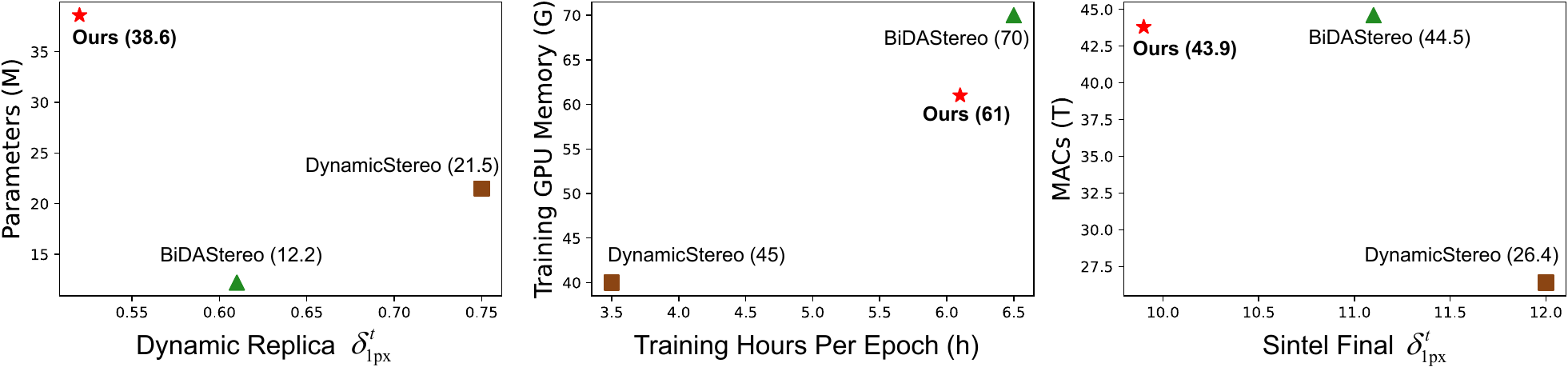} 
    \caption{(a) $\delta^{t}_{1px}$ on DR \textit{vs.} parameters. (b) Training GPU memory at $320\times 512$ \textit{vs.} Training hours per epoch. (c) $\delta^{t}_{1px}$ on Sintel \textit{vs.} MACs (20 frames $\times$ 768 $\times$ 1024).} 
    \label{sec4:params}
\end{figure*}

As illustrated in Fig.~\ref{sec4:params}, we conduct a comprehensive comparison of the competing methods across three critical metrics: model size (parameters), training GPU memory consumption, and computational complexity (multiply–accumulate operations, MACs). Our proposed method achieves an optimal trade-off among these efficiency criteria while simultaneously delivering the lowest error rate. Notably, compared to the previous state-of-the-art approach, BiDAStereo~\cite{jing2024match}, our method demonstrates a significant performance improvement while maintaining comparable computational costs. The advantage of enhanced accuracy and superior efficiency makes our approach particularly suitable for real-world applications.

\subsection{Memory Reference}
~\label{sec:c}
Here, we visualize the memory aggregation process (Section 3) by showing the candidate frames, some of the selected reference frames, and the corresponding aggregation weights. As illustrated in Figure~\ref{sec4:attention_vis} we observe semantically meaningful
regions to be focused.

\section{Limitations and Future}
~\label{sec:d}
While our method advances the state of dynamic scene modeling, it shares a common limitation with existing approaches: the inability to proactively distinguish between dynamic and static regions, which is crucial for maintaining temporal consistency.
Also, our method occasionally in textureless areas (e.g., blank walls) or transparent surfaces (e.g., glass), where current techniques, including ours, may produce inconsistencies.
To address these limitations, we plan to pursue two key directions: (1) integrating high-quality memory cues to improve scene understanding and consistency, and (2) developing a lightweight variant of our model for resource-constrained applications~\cite{huang2023distributed,huang2024fedmef,huang2025fedrts,huang2025quaff,huang2025tequila,wang2025interventional}. Looking forward, we aim to create a comprehensive model zoo featuring both full-capacity and efficient versions of our approach, facilitating adoption across different hardware scenarios.

\end{document}